\documentclass[journal,twoside]{IEEEtran}

\usepackage{amsmath,amsfonts}
\usepackage{algorithmic}
\usepackage{algorithm}
\usepackage{array}
\usepackage[caption=false,font=footnotesize,labelfont=rm,textfont=rm]{subfig}
\usepackage{textcomp}
\usepackage{stfloats}
\usepackage{url}
\usepackage{verbatim}
\usepackage{graphicx}
\usepackage{cite}
\usepackage{enumitem}
\usepackage[breaklinks,colorlinks, linkcolor=red, citecolor=blue, urlcolor=black]{hyperref}

\usepackage[utf8]{inputenc} 
\usepackage[T1]{fontenc}    
\usepackage{hyperref}       
\usepackage{url}            
\usepackage{booktabs}       
\usepackage{amsfonts}       
\usepackage{nicefrac}       
\usepackage{microtype}      
\usepackage{xcolor}         
\usepackage{multirow}
\usepackage{graphicx}
\usepackage{epstopdf}
\usepackage{amsmath}
\usepackage{amssymb}
\usepackage{utfsym}
\usepackage{pifont}
\usepackage{bm}
\usepackage{float}
\usepackage{subfig}
\usepackage{wrapfig}
\newcommand{\ie}{{\emph{i.e.}}}
\newcommand{\eg}{{\emph{e.g.}}}
\newcommand{\etal}{{\emph{et al.}}}

\newcommand{\cmark}{\ding{51}}%
\newcommand{\xmark}{\ding{55}}

\DeclareMathOperator*{\argmin}{arg\,min}

\usepackage{xcolor}
\usepackage{colortbl}

\begin{document}

\title{Semantics-Oriented Multitask Learning for DeepFake Detection: A Joint Embedding Approach}

\author{Mian~Zou,
        Baosheng~Yu,
        Yibing~Zhan,~\IEEEmembership{Member,~IEEE,}
        Siwei~Lyu,~\IEEEmembership{Fellow,~IEEE,}
        and~Kede~Ma,~\IEEEmembership{Senior Member,~IEEE}
\thanks{Mian Zou and Kede Ma are with the Department of Computer Science, City University of Hong Kong, Kowloon, Hong Kong (e-mail: mianzou2-c@my.cityu.edu.hk; kede.ma@cityu.edu.hk).}
\thanks{Baosheng Yu is with the Lee Kong Chian School of Medicine, Nanyang Technological University, Singapore (e-mail: baosheng.yu@ntu.edu.sg).}
\thanks{Yibing Zhan is with Yunnan United Vision Technology Co., Ltd, Yunnan, China (e-mail: zybjy@mail.ustc.edu.cn).}
\thanks{Siwei Lyu is with the Department of Computer Science and Engineering, University at Buffalo, State University of New York, Buffalo, NY USA (e-mail: siweilyu@buffalo.edu).}
\thanks{\emph{Corresponding author: Kede~Ma}.}
}

\markboth{IEEE Transactions on Circuits and Systems for Video Technology}%
{Zou \MakeLowercase{\textit{et al.}}: Semantics-Oriented Joint Embedding DeepFake Detection}


\maketitle

\begin{abstract}
In recent years, the multimedia forensics and security community has seen remarkable progress in multitask learning for DeepFake (\ie, face forgery) detection. The prevailing approach has been to frame DeepFake detection as a binary classification problem augmented by manipulation-oriented auxiliary tasks. This scheme focuses on learning features specific to face manipulations with limited generalizability. 
In this paper, we delve deeper into semantics-oriented multitask learning for DeepFake detection, capturing the relationships among face semantics via joint embedding. We first propose an automated dataset expansion technique that broadens current face forgery datasets to support semantics-oriented DeepFake detection tasks at both the global face attribute and local face region levels. Furthermore, we resort to the joint embedding of face images and labels (depicted by text descriptions) for prediction. This approach eliminates the need for manually setting task-agnostic and task-specific parameters, which is typically required when predicting multiple labels directly from images. In addition,
we employ bi-level optimization to dynamically balance the fidelity loss weightings of various tasks, making the training process fully automated. Extensive experiments on six DeepFake datasets show that our method improves the generalizability of DeepFake detection and renders some degree of model interpretation by providing human-understandable explanations.

\end{abstract}


\section{Introduction}~\label{sec:intro}
\IEEEPARstart{T}{he} emergence of deep learning~\cite{faceswap,karras2020analyzing,song2019generative,thies2019deferred} has greatly simplified and automated the production of realistic counterfeit face images and videos, commonly referred to as DeepFakes. This seriously threatens the authenticity and integrity of digital visual media. In response, a wide array of DeepFake detection methods have been developed to efficiently screen manipulated face imagery~\cite{juefei2022countering, wang2024deepfake}.

Early DeepFake detectors~\cite{matern2019exploiting,mccloskey2019detecting,durall2019unmasking} were largely shaped by established photo forensics techniques~\cite{farid2009image}, through the analysis of statistical anomalies~\cite{popescu2005exposing}, visual artifacts~\cite{johnson2006exposing,lyu2010estimating,lyu2014exposing}, and physical inconsistencies~\cite{johnson2007exposing,kee2014exposing,o2012exposing}.
The advent of deep learning has spurred the development of numerous learning-based detection methods~\cite{zhou2017two, afchar2018mesonet, rossler2019faceforensics, wang2019cnngenerated, Nguyen2019multi, li2020face, Cao_2022_CVPR, chen2022self, Dong_2023_CVPR, Yan_2023_ICCV, zhao2021multi, haliassos2021lips, luo2021generalizing, shiohara2022detecting, dong2022protecting, sun2022dual, wang2023cvpr_SFDG, sun2023general, yang2023masked, zhang2024temporal, Luo2024tifs_CFM}, operating as binary classifiers to predict whether a test image is real or fake. 
This formulation can be augmented by various auxiliary tasks, such as manipulation type identification~\cite{he2021forgerynet, Sun_2023_ICCV}, manipulation parameter estimation~\cite{wang2019detecting}, blending boundary detection~\cite{li2020face}, and face reconstruction~\cite{Cao_2022_CVPR,Yan_2023_ICCV}), leading to a multitask learning paradigm~\cite{Nguyen2019multi, li2020face, pcl, Cao_2022_CVPR, chen2022self, Dong_2023_CVPR, Yan_2023_ICCV, zhao2021multi, zhang2024adaptive}. Here, the underlying assumption is that the primary task will likely gain from auxiliary tasks through knowledge transfer. 
However, most auxiliary tasks are specific to manipulations, which may hinder the generalizability of DeepFake detectors. Recently, Zou \etal~\cite{zou2024semantic} introduced a semantics-oriented multitask learning paradigm for DeepFake detection. They argued that face forgery stems from computational methods that modify semantic face attributes beyond the thresholds of human discrimination. Although conceptually and computationally appealing, training semantics-oriented DeepFake detectors would necessitate considerable human annotations to specify manipulation degree parameters and semantic label relationships~\cite{zou2024semantic}. Moreover, the allocation of task-agnostic and task-specific parameters is optimized manually, which is bound to be suboptimal in facilitating knowledge transfer among tasks.

\begin{figure}[!t]
  \centering
  \includegraphics[width=0.9\linewidth]{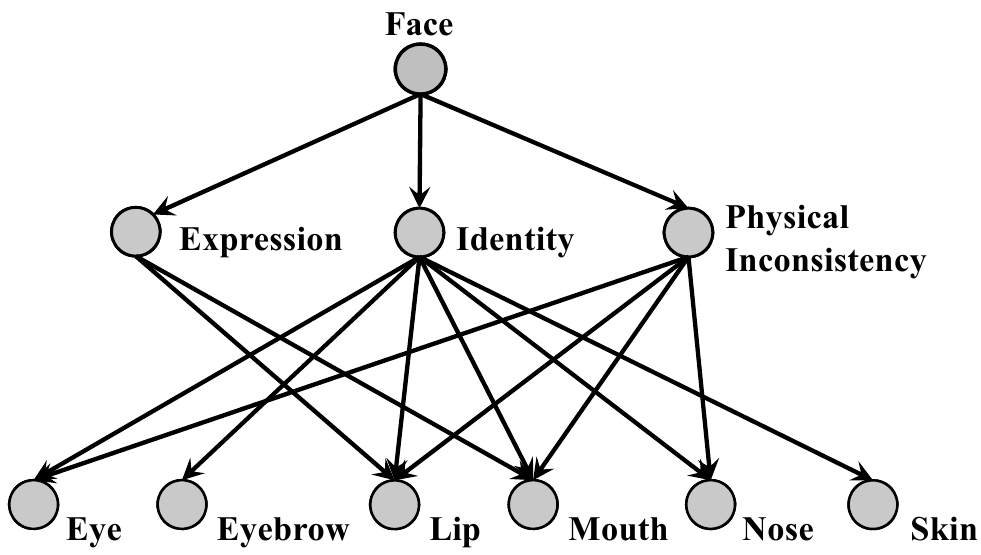}
  \caption{Illustration of the label hierarchy in our expanded FF++ dataset~\cite{rossler2019faceforensics}. The hierarchy organizes manipulation types into three  nodes: \texttt{identity} (face-swapping via Deepfakes~\cite{faceswap} and FaceSwap~\cite{faceswap_Kowalski}), \texttt{expression} (mouth editing via Face2Face~\cite{thies2016face2face} and NeuralTextures~\cite{thies2019deferred}), and \texttt{physical\_inconsistency} (local face editing via data augmentations). These nodes connect to six leaf nodes representing six distinct local face regions. Edges denote manipulation relationships: \texttt{identity} affects all regions, \texttt{expression} targets \texttt{lip} and \texttt{mouth}, while \texttt{physical\_inconsistency} modifies \texttt{eye}, \texttt{lip}, \texttt{mouth}, and \texttt{nose}. }
  \label{fig: graph}
\end{figure}

In this paper, we further pursue the promising direction of semantics-oriented multitask learning for DeepFake detection with two important improvements. First, we present a dataset expansion technique that automatically enriches the binary labeling of existing DeepFake datasets to semantic hierarchical labeling while increasing their sizes through data augmentation (see Fig.~\ref{fig: graph}). A crucial element is to incorporate the \texttt{physical\_inconsistency} node that encompasses local face region manipulations, without manually identifying interdependent changes in other global face attributes (if any). 
Second, unlike discriminative learning that directly predicts labels from images, we take inspiration from the recent advances in vision-language correspondence~\cite{radford2021CLIP} and encode both the input image and label hierarchy (represented by text templates) in a common feature space. This enables us to view model parameter sharing/splitting in multitask learning from the perspective of model capacity allocation. More specifically, we share the image encoder parameters across all tasks, while the capacity allocated to each individual task is end-to-end optimized. 
In addition,  we adopt bi-level optimization~\cite{dempe2002foundations} to prioritize the primary task by minimizing its fidelity loss~\cite{tsai2007frank} with respect to the loss weighting vector. The parameters of the DeepFake detector are trained by minimizing the weighted sum of fidelity losses of individual tasks.

Extensive experiments on six datasets~\cite{rossler2019faceforensics, Li_2020_CVPR, li2020celeb, jiang2020deeperforensics, Dolhansky2020deepfake, zou2024semantic} demonstrate that our {\em Semantics-oriented Joint Embedding DeepFake Detector} (SJEDD) consistently improves the cross-dataset, cross-manipulation, and cross-attribute detection performance, compared to state-of-the-art DeepFake detectors. 
In summary, our contributions are threefold:
\begin{itemize}
	 \item an automated dataset expansion technique that enriches the binary labeling of DeepFake datasets,
      \item a semantics-oriented joint embedding method for DeepFake detection, coupled with bi-level optimization to fully automate the training process, and
      \item an experimental demonstration that the proposed SJEDD offers better generalizability and interpretability.
\end{itemize}

\section{Related Work}\label{sec: related_work}
In this section, we provide a concise review of DeepFake detectors, emphasizing the multitask learning paradigm. We also discuss joint embedding architectures in the broader context of machine learning.

\subsection{DeepFake Detectors}
\noindent\textbf{Non-learning-based detectors} rely on extensive domain expertise. The first type of knowledge involves \textit{mathematical} models of the image source, which summarizes the essential elements of what constitutes an authentic image~\cite{popescu2005exposing}.
The second type of knowledge involves \textit{computational} models of the digital photography pipeline, along which each step can be exploited to expose forgery~\cite{farid2016photo_forensics}, including the lens distortion and flare~\cite{johnson2006exposing,lyu2010estimating}, color filter array~\cite{popescu2005exposing2}, sensor noise \cite{lyu2014exposing,wang2023noise}, pixel response non-uniformity~\cite{lukas2006digital}, camera response function \cite{lin2005detecting}, and file storage format~\cite{kee2011digital}. The third type of knowledge involves \textit{physical} and \textit{geometric} models of our three-dimensional world, especially those that capture the interactions of light, optics, and objects. Face forgery can be exposed by detecting physical and geometric
inconsistencies, such as those found in illumination \cite{johnson2007exposing}, shadow \cite{kee2014exposing}, reflection \cite{o2012exposing}, and motion \cite{conotter2011exposing}. Another line of research that has received significant attention involves detecting abnormal physiological reactions from eye blinking~\cite{Li2018in}, pupil shape~\cite{guo2022icassp_eyes}, head pose~\cite{yang2019exposing}, and corneal specularity~\cite{hu2021exposing}. Nevertheless, these physiological features may not sufficiently capture the nuances needed to identify face forgery. 

\noindent\textbf{Learning-based methods} learn to detect DeepFake images and videos from data.
Afchar \etal~\cite{afchar2018mesonet} proposed to examine DeepFake videos at the mesoscopic level. Zhou \etal~\cite{zhou2017two} designed a two-stream network to jointly analyze camera characteristics and visual artifacts. Li and Lyu \cite{Li_2019_CVPR_Workshops} aimed to detect affine warping artifacts, while Li \etal~\cite{li2020face} and Nguyen \etal~\cite{nguyen2024laa} focused on locating image blending boundaries. 
Dong \etal~\cite{Dong_2023_CVPR} revealed a shortcut termed ``implicit identity leakage'' and proposed an identity-unaware DeepFake detector. Gu \etal~\cite{gu2021spatiotemporal} and  Zheng \etal~\cite{Zheng_2021_ICCV} examined spatiotemporal inconsistencies at the signal level, while Haliassos~\etal~\cite{haliassos2021lips} did so at the semantic level through lipreading pretraining. Xu~\etal~\cite{Xu_2023_ICCV} proposed a thumbnail layout to trade spatial modeling for temporal modeling. 
Additionally, complex architectures such as capsule networks~\cite{nguyen2019capsule} and vision Transformers~\cite{lu2023detection, dong2022protecting,wang2023deep}, along with advanced training strategies such as adversarial training~\cite{chen2022self, li2022artifacts}, test-time training~\cite{chen2022ost}, graph learning~\cite{wang2023cvpr_SFDG, yang2023masked}, contrastive learning~\cite{sun2022dual, Luo2024tifs_CFM}, and one-class classification~\cite{Feng_2023_CVPR}, have also been explored for DeepFake detection. Generally, learning-based methods typically overfit to training-specific artifacts~\cite{zhou2017two, Li_2019_CVPR_Workshops, Dong_2023_CVPR, wang2023noise, li2020face, zhang2023pyrf}. 

\noindent\textbf{Multitask learning-based methods} have emerged as a promising offshoot of learning-based methods for DeepFake detection in the past five years. By leveraging auxiliary supervisory signals during training, these methods tend to offer improved generalizability compared to single-task approaches.
Notable auxiliary tasks include manipulation localization~\cite{Nguyen2019multi, chen2022self} (with blending boundary detection~\cite{li2020face} as a special case), manipulation classification and attribution~\cite{he2021forgerynet,Sun_2023_ICCV}, manipulation parameter estimation~\cite{wang2019detecting}, face image reconstruction~\cite{Cao_2022_CVPR,Yan_2023_ICCV}, and tasks that promote robust and discriminative feature learning~\cite{zhao2021multi}. Most auxiliary tasks for DeepFake detection are manipulation-oriented, which may overfit detectors to manipulation-specific patterns and limit their effectiveness against sophisticated, novel forgery.
To address this limitation, Zou~\etal~\cite{zou2024semantic} pioneered a paradigm shift from manipulation-oriented to semantics-oriented detection. Their approach emphasizes the joint learning of high-level representations of global face attributes and low-level representations of local face regions, thus enhancing detection generalizability.
In this paper, we further advance this promising paradigm.
From the \textit{data} perspective, we introduce an automated dataset expansion technique that not only enlarges existing DeepFake datasets but also enriches them with a collection of labels organized in a semantic hierarchical graph (see Fig.~\ref{fig: graph}). From the \textit{model} perspective, we leverage joint embedding instead of discriminative learning to automate model parameter sharing/splitting and to facilitate task relationship modeling.
Concurrently, Sun~\etal~\cite{sun2023general} proposed VLFFD, which also implemented DeepFake detection via joint embedding and multitask learning. 
At a high level, SJEDD employs semantics-oriented multitask learning, while VLFFD adopts a manipulation-oriented approach, centered on testing the feasibility of CLIP~\cite{radford2021CLIP} for DeepFake detection. This distinction drives four key computational differences. First, SJEDD uses probabilistic modeling on a semantic hierarchical graph, while VLFFD relies on the standard CLIP architecture. Second, SJEDD prioritizes semantics-oriented tasks with text prompts, whereas VLFFD focuses on identifying manipulation artifacts. Third, SJEDD applies principled bi-level optimization for automated multitask learning, contrasting with the heuristic tuning of task-specific hyperparameters in VLFFD. 

\begin{figure*}[!ht]
    \centering
  \includegraphics[width=\linewidth]{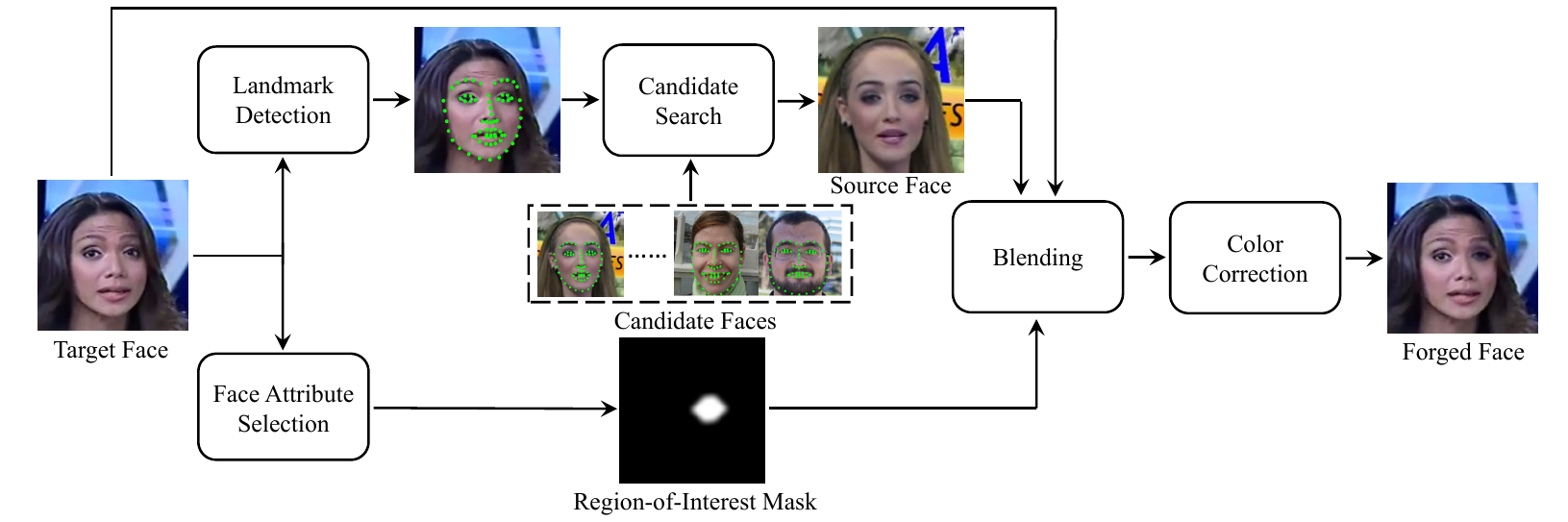}
  \caption{Pipeline for local face region manipulation.  When searching for the best candidate as the source image, we minimize the Euclidean distance between the $68$ detected landmarks~\cite{li2020face} of the target and candidate face images (excluding those with the same identity). 
  }
  \label{fig: tdg_phys}
\end{figure*}

\subsection{Joint Embedding Architectures}
Joint embedding architectures~\cite{lecun2022path} aim to learn embeddings that are close for compatible inputs (\eg, an input image and its target label) and distant for incompatible inputs, as opposed to discriminative approaches that directly predict labels from inputs. 
This idea can be traced back to Siamese networks~\cite{bromley1993signature}, initially proposed for signature verification and later adapted to other computer vision tasks like face verification~\cite{chopra2005learning} and recognition~\cite{taigman2014deepface}, and video feature learning~\cite{taylor2011learning}.
With the rise of self-supervised learning, a thorough exploration of joint embedding architectures is conducted, contrastively (\eg, MoCo~\cite{he2020momentum} and SimCLR~\cite{chen2020simple}) and non-contrastively (\eg, Barlow Twins~\cite{zbontar2021barlow} and I-JEPA~\cite{assran2023self}). The more recent vision-language models~\cite{jia2021scaling,radford2021CLIP} are also manifestations of joint embedding architectures and are quickly adapted to multiple downstream tasks such as image quality assessment~\cite{zhang2023liqe}, realistic image enhancement~\cite{liang2023iterative}, image-text retrieval~\cite{li2021align},  and multimodal fake news detection~\cite{zhou2022multimodal}.

Recent DeepFake detectors have also leveraged vision-language joint embedding~\cite{wu2023generalizable,khan2024clipping, sun2023general, lai2024gmdf, han2024fcg, lin2024Repdfd}. While these methods directly utilize cosine similarity scores for detection, the proposed SJEDD maps these scores into a joint probability distribution over a customized semantic label hierarchy. Our design explicitly integrates label relationships and unifies low-level visual features with high-level semantic cues, thus facilitating semantics-oriented DeepFake detection.

\section{Semantics-Oriented Dataset Expansion: A Demonstration on FF++}\label{subsec:tdg}
In this section, we describe our automated dataset expansion technique, consisting of two steps: data augmentation and data relabeling. Data augmentation generates new images by manipulating local face regions. Data relabeling automatically assigns DeepFake images with a set of labels through semantic contextualization of the adopted face manipulations~\cite{zou2024semantic}. 
To demonstrate the proposed dataset expansion technique, we apply it to expand the widely used FF++ dataset~\cite{rossler2019faceforensics}.

\subsection{Data Augmentation}
We augment FF++ by manipulating four local face regions, \ie, \texttt{eye}, \texttt{lip}, \texttt{mouth}, and \texttt{nose}, as illustrated in Fig.~\ref{fig: tdg_phys}. Given a real face as the target, we initially search for the best source face (real or fake) by minimizing the Euclidean distance between their corresponding $68$ detected landmarks~\cite{bulat2017far} while excluding faces with the same identity~\cite{li2020face}. Next, we linearly blend the local face region from the source, delineated by a Gaussian-blurred mask, into the target image, followed by color correction~\cite{reinhard2001color}. 
In our implementation, candidate face images are sourced from the Deepfakes~\cite{faceswap} and FaceSwap~\cite{faceswap_Kowalski} subsets of FF++. For each of the four local face regions, the number of augmented images matches that of forged images by each face manipulation method in the original FF++.

\begin{table*}[]
\renewcommand{\arraystretch}{1.2}
\caption{Semantical hierarchical labeling of the expanded FF++ dataset: ``1'' indicates that the face attribute (or region) has been modified, ``0'' means no modification, and ``N/A'' denotes that the underlying label  is unobserved}
\label{tab: so-data_expand}
\small
\centering
\resizebox{\linewidth}{!}{
\begin{tabular}{llcccccccccc}
\toprule
                                   & Manipulation / Region      & \texttt{Face} & \texttt{Expression} & \texttt{Identity} & \begin{tabular}[c]{@{}c@{}}\texttt{Physical}\\ \texttt{Inconsistency}\end{tabular} & \texttt{Eye} & \texttt{Eyebrow} & \texttt{Lip} & \texttt{Mouth} & \texttt{Nose} & \texttt{Skin} \\
\hline
\noalign{\smallskip}
\multirow{3}{*}{Data Augmentation} & Eye & 1    & N/A        & 0        & 1                       & 1   & 0       & 0   & 0     & 0    & 0    \\
                                   & Lip and Mouth & 1    & N/A        & 0        & 1                       & 0   & 0       & 1   & 1     & 0    & 0    \\
                                   & Nose & 1    & 0          & 0        & 1                       & 0   & 0       & 0   & 0     & 1    & 0  \\
\hline  
\noalign{\smallskip}
\multirow{4}{*}{Data Relableling}  & Deepfakes~\cite{faceswap}          & 1    & N/A          & 1        & 1                       & 1   & 1       & 1   & 1     & 1    & 1    \\
                                   & Face2Face~\cite{thies2016face2face}          & 1    & 1          & 0        & 1                       & 0   & 0       & 1   & 1     & 0    & 0    \\
                                   & FaceSwap~\cite{faceswap_Kowalski}           & 1    & N/A          & 1        & 1                       & 1   & 1       & 1   & 1     & 1    & 1    \\
                                   & NeuralTextures~\cite{thies2019deferred}     & 1    & 1          & 0        & 1                       & 0   & 0       & 1   & 1     & 0    & 0    \\
\bottomrule  
\end{tabular}
}
\end{table*}

\subsection{Data Relabeling}~\label{subsec: dataset_expansion_relabel}
Inspired by~\cite{zou2024semantic}, we first construct a semantic hierarchical graph to contextualize the four face manipulations of FF++: Deepfakes~\cite{faceswap}, FaceSwap~\cite{faceswap_Kowalski}, Face2Face~\cite{thies2016face2face}, and NeuralTextures~\cite{thies2019deferred}. The former two swap faces between images to achieve \textit{identity} manipulation, while the latter two change the \textit{expression} of target faces through manipulation of the mouth region. As a result, we instantiate two global face attribute nodes: \texttt{expression} and \texttt{identity}. We also include six leaf nodes, representing nonoverlapping local face regions: \texttt{eye}, \texttt{eyebrow}, \texttt{lip}, \texttt{mouth}, \texttt{nose}, and \texttt{skin}. For the two identity manipulations based on face swapping, all local face regions are assumed to be altered. In contrast, for the two expression manipulations, only \texttt{lip} and \texttt{mouth} regions are affected. To accommodate local face manipulations in the graph, we introduce an additional global face attribute node, \texttt{physical\_inconsistency}, which signifies images where local face regions are replaced by the corresponding regions from source images, resulting in physical incompatibility.

\subsection{Discussion}~\label{subsec: dataset_expansion_discuss}
The proposed dataset expansion technique allows some labels in the semantic hierarchical graph to remain unobserved, which improves labeling reliability. For instance, in the case of face swapping, it is unclear whether the \texttt{expression} has been altered, so we leave this label unobserved. Meanwhile, in the expanded dataset, a single face manipulation can simultaneously alter multiple face attributes  (\eg,  \texttt{identity} and \texttt{physical\_consistency} by FaceSwap~\cite{faceswap_Kowalski}), while different manipulations can also modify the same attribute (\eg, \texttt{identity} by Deepfakes~\cite{faceswap} and FaceSwap). As such, our technique emphasizes a semantics-oriented rather than manipulation-oriented contextualization of face manipulations akin to FFSC~\cite{zou2024semantic} and achieves so without expensive human labeling. 
Table~\ref{tab: so-data_expand} presents the list of face manipulations in the expanded FF++ dataset, along with the corresponding modified face attributes and regions.

\section{Proposed Method: SJEDD}\label{sec: method}
\begin{figure*}[t]
    \centering
  \includegraphics[width=0.9\linewidth]{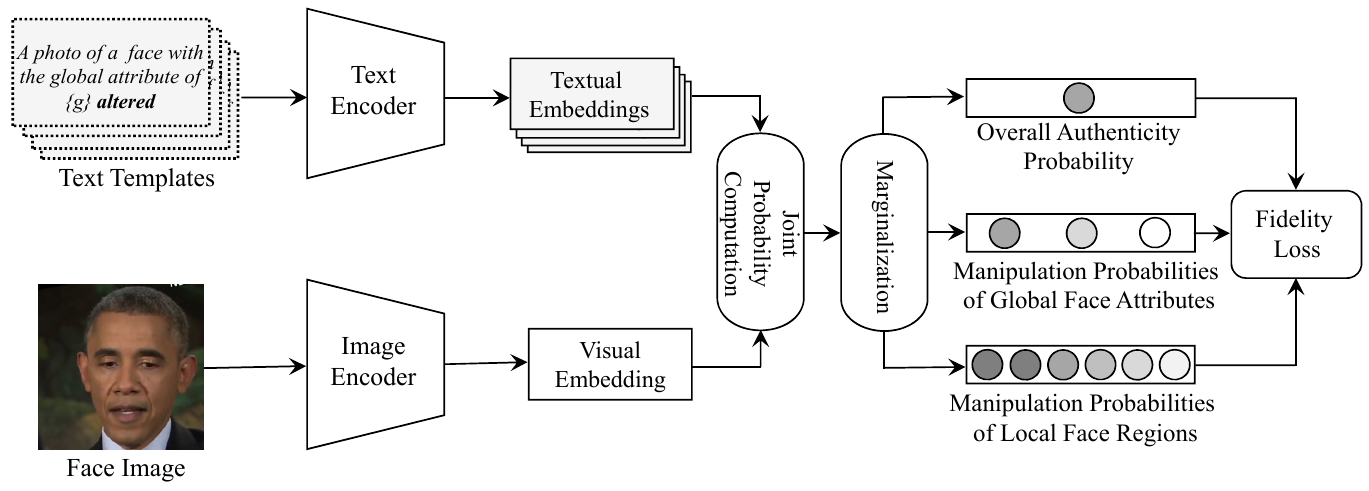}
  \caption{System diagram of SJEDD.}
  \label{fig: framework}
\end{figure*}

In this section, we describe the proposed DeepFake detector SJEDD in detail, including 1) general problem formulation, 2) joint embedding for DeepFake detection, and 3) bi-level optimization for parameter estimation. 
The system diagram of SJEDD is shown in Fig.~\ref{fig: framework}.

\subsection{Problem Formulation}
Given an input face image $\bm x\in\mathbb{R}^{H\times W\times 3}$, we assign to it a vector of binary labels $\bm y = \{0,1\}^{C}$, corresponding to one primary and $C-1$ auxiliary tasks. Specifically, $y_1=1$ signifies a forged face, and $y_1=0$ means the opposite. For $i>1$, $y_i=1$ indicates that the $i$-th face attribute/region is fake, and $y_i=0$ otherwise. Next, we define an unnormalized probability distribution over the label hierarchy:
\begin{align}
    \label{eq:plf_joint}
    \begin{split}
        \tilde{p}(\bm y|\bm x) =& \prod_i e^{s_i(\bm x)\mathbb{I}[y_i=1]}\prod_{
           j, i\in \mathcal{P}_j
         } \mathbb{I}\left[\left(\sum_{i} y_i,y_j\right)\ne (0,1)\right]\\
        &\prod  _{
           i , j\in \mathcal{C}_i
         } \mathbb{I}\left[\left( y_i,\sum_{j}y_j\right)\ne (1,0)\right],
    \end{split}
\end{align}
where $s_i(\bm x)$ represents the raw score of the detector for the $i$-th label, and $\mathbb{I}[\cdot]$ is an indicator function. Eq.~\eqref{eq:plf_joint} emphasizes two label relations: 1) if the $j$-th node is activated (\ie, $y_j = 1$), at least one of its parent nodes, collectively indexed by $\mathcal{P}_j$, must also be activated; and 2) if the $i$-th node is activated, at least one of its child nodes, indexed by $\mathcal{C}_i$, must also be activated.
We then normalize Eq.~\eqref{eq:plf_joint} to obtain the joint probability: 
\begin{align}\label{eq:joint}
    \hat{p}(\bm y|\bm x)=\frac{\tilde{p}(\bm y|\bm x)}{Z(\bm x)}, \,\textrm{where}\, Z(\bm x) = \sum_{\bm y'} \tilde{p}(\bm y'|\bm x).
\end{align}
Due to the small scale of our label hierarchy, $Z(\bm x)$ can be easily computed using matrix multiplication. We further compute the marginal probability of a given label $y_i$ by marginalizing the joint probability over all other labels:
\begin{align}\label{eq: plf_marginal}
    \hat{p}(y_i|\bm x) = \sum_{\bm y \setminus y_{i}}\hat{p}(\bm y \vert \bm x),
\end{align}
where $\setminus$ denotes the set difference operation.

\subsection{Joint Embedding}\label{subsec: joint_embedding}
We now describe the text templates to encode the label hierarchy and implement joint embedding using vision-language correspondence to estimate the raw scores for all labels.

\subsubsection{Text Templates}~\label{subsec: text_templates}
We create the text templates based on face attributes, explicitly incorporating semantic face priors into the joint embedding space.
For the root node \texttt{face}, the text template is defined as ``\textit{a photo of a fake face}.'' At the global face attribute level, we use ``\textit{A photo of a face with the global attribute of $\{g\}$ altered},'' where $g$ belongs to the set of global face attributes \{\texttt{expression}, \texttt{identity}, \texttt{physical\_inconsistency}\}. At the local face region level, we adopt ``\textit{A photo of a face with the local $\{l\}$ region altered},'' where $l$ is in the set of local face regions \{\texttt{eye}, \texttt{eyebrow}, \texttt{lip}, \texttt{mouth}, \texttt{nose}, \texttt{skin}\}.
In total, we have ten candidate text templates.

\subsubsection{Vision-Language Correspondence}
We construct two encoders: an image encoder $\bm f_{\bm\phi}:\mathbb{R}^{H\times W\times 3} \mapsto \mathbb{R}^{N\times 1}$, parameterized by $\bm\phi$, which computes the visual embedding $\bm f_{\bm\phi}(\bm x)$ from an input face image $\bm x$, and a text encoder $\bm g_{\bm \varphi}:\mathcal{T} \mapsto \mathbb{R}^{N\times 1}$, parameterized by $\bm \varphi$, which produces a set of textual embeddings $\{\bm g_{\bm \varphi}(\bm t_i)\}_{i=1}^C$. Here, $\bm t_i$ denotes the $i$-th text template, and $\mathcal{T}$ encompasses all candidate text templates.
We estimate the raw score in Eq.~\eqref{eq:plf_joint} by
\begin{align}
    \label{eq:unnorm_similarity}
    s_i(\bm x)= \frac{\langle\bm f_{\bm \phi}(\bm x),\bm g_{\bm \varphi}(\bm t_i)\rangle}{\tau},
\end{align}
where $\langle\cdot,\cdot\rangle$ denotes the inner product, and $\tau$ is a pre-determined temperature parameter.
The raw scores are fed into the joint distribution defined in Eq.~\eqref{eq:plf_joint}, which is then normalized and marginalized to compute the manipulation probability for each node in Fig.~\ref{fig: graph}. Higher probabilities indicate that an alteration has likely occurred.

The joint embedding approach eliminates the need for manual splitting of task-agnostic and task-specific layers. Through end-to-end optimization, it learns to allocate model capacity for each task automatically.

\subsection{Bi-level Optimization}
Our goal is to optimize the joint embedding function, parameterized by $\bm \phi$ and $\bm \varphi$, which outputs the raw scores for all labels, based on which their marginal probabilities $\hat{p}(y_i|\bm x;\bm \phi, \bm \varphi)$ can be computed.
Given a training minibatch $\mathcal{B}_\mathrm{tr} = \{\bm x^{(k)}, \bm y^{(k)}, \mathcal{Z}^{(k)}\}_{k=1}^K$, where $\bm y ^{(k)} = \{0,1\}^{C}$ is the complete ground-truth label vector and $\mathcal{Z}^{(k)}\subseteq \{1,\ldots, C\}$ is the index set of observed labels, we minimize a linear weighted sum of losses for each observed label:
\begin{align}
    \label{eq:overallloss}
        \ell(\mathcal{B}_\mathrm{tr};\bm \phi, \bm \varphi)
        =&\frac{1}{K}\sum_{k=1}^{K} \sum_{i \in \mathcal{Z}^{(k)}} \lambda_{i} \ell\left(\bm x^{(k)}, y^{(k)}_i; \bm \phi, \bm \varphi\right), 
\end{align}
where $\bm \lambda = [\lambda_1,
\ldots,\lambda_{C}]^\intercal$ is the loss weighting vector to be optimized along with the model parameters $\bm \phi$ and $\bm \varphi$.

While the de facto cross-entropy loss can be used to implement $\ell(\cdot)$ in Eq.~\eqref{eq:overallloss} for parameter estimation, it is unbounded from above~\cite{tsai2007frank}. This may result in excessive penalties for challenging training examples, thus hindering model capacity allocation and loss weighting adjustment in multitask learning. To address this, we use the fidelity loss~\cite{tsai2007frank}:
\begin{align}
    \label{eq: loss_b}
    \ell(\bm x, y_i;\bm \phi, \bm \varphi) =& 1 - \sqrt{p(y_i|\bm x)\hat{p}(y_i|\bm x)} \nonumber  \\
                               &- \sqrt{(1 - p(y_i|\bm x))(1-\hat{p}(y_i|\bm x))}, 
\end{align}
where $p(y_i|\bm x)$ denotes the $i$-th ground-truth label. 
 
To automatically adjust the loss weighting vector $\bm \lambda$ and to prioritize the primary DeepFake detection task, we employ bi-level optimization~\cite{dempe2002foundations}. At the outer level, we minimize the primary task objective with respect to $\bm \lambda$ on the validation minibatch $\mathcal{B}_\mathrm{val}$. At the inner level, we minimize the linearly weighted loss with respect to $\bm \phi$ and $\bm \varphi$ on the training minibatch $\mathcal{B}_\mathrm{tr}$, giving rise to
\begin{align}
\begin{split}
\label{eq:auto-weighting}
\min_{\bm \lambda} \frac{1}{\vert\mathcal{B}_\mathrm{val}\vert }  &\sum_{\bm x\in \mathcal{B}_{\mathrm{val}}} \ell\left (\bm  x, y_1; {\bm \phi}^\star, {\bm \varphi}^\star \right)  \\
\text{s.t.}\quad {\bm \phi}^\star, {\bm \varphi}^\star &= \argmin_{\bm \phi, \bm \varphi}  \ell(\mathcal{B}_\mathrm{tr};\bm \phi, \bm \varphi),
\end{split}
\end{align}
where $\vert\cdot\vert$ denotes the cardinality of a set. As suggested in~\cite{zou2024semantic}, we share loss weightings for tasks within the same level. That is, we learn to adjust three $\lambda_i$'s: one for the primary task, one for tasks at the global face attribute level, and one for tasks at the local face region level. 
We sample training and validation data as different minibatches in the same training dataset. We utilize the finite difference method as an efficient approximator to solve Problem \eqref{eq:auto-weighting}.

\begin{table*}[!t]
\renewcommand{\arraystretch}{1.1}
\caption{AUC results of the proposed SJEDD against existing DeepFake detectors in the cross-dataset setting. 
All models are trained using the (respectively augmented) training set of FF++. We retrain CNND, Face X-ray, MADD, FRDM, SLADD, DCL, and SO-ViT-B to match the performance reported in their original papers. For SFDG, VLFFD, CCDFM, and Zhang24, we directly copy the results from their original papers due to the lack of publicly available implementations. The last column shows the mean AUC numbers, including/excluding the FF++ test set. The best results are highlighted in bold
}
\label{tab:cross_dataset_test}
\centering
\resizebox{\linewidth}{!}{
\begin{tabular}{lllccccccc}
\toprule
 Method  & Venue & Backbone & FF++~\cite{rossler2019faceforensics} & CDF~\cite{li2020celeb} & FSh~\cite{Li_2020_CVPR} & DF-1.0~\cite{jiang2020deeperforensics} & DFDC~\cite{Dolhansky2020deepfake} & FFSC~\cite{zou2024semantic} & Mean AUC \\
\hline
\noalign{\smallskip}
CNND~\cite{wang2019cnngenerated} & CVPR2020 & ResNet-50 & 98.13          & 75.60          & 65.70          & 74.40         & 72.10 & 61.58 & 74.59 / 69.88 \\
Face X-ray~\cite{li2020face} & CVPR2020 & HRNet  & 98.37          & 80.43          & 92.80          & 86.80          & 65.50 & 76.57 & 83.41 / 80.42 \\
MADD~\cite{zhao2021multi} & CVPR2021 & EfficientNet-b4  & 98.97          & 77.44          & 97.17          & 66.58          & 67.94 & 80.91 & 81.50 / 78.01 \\
Lip-Forensics~\cite{haliassos2021lips} & CVPR2021 & ResNet-18  & 99.70 & 82.40          & 97.10          & \textbf{97.60}          & 73.50 & \textbf{---} & \textbf{---} \\
FRDM~\cite{luo2021generalizing} & CVPR2021 & Xception  & 98.36    &  79.40   &  97.48       &      73.80     & 79.70 & 75.61 & 84.06 / 81.20 \\ 
RECCE~\cite{Cao_2022_CVPR} &  CVPR2022 & Xception  & 99.32        & 68.71          & 70.58          & 74.10          & 69.06 & 77.86 & 76.61 / 72.06 \\
SBI~\cite{shiohara2022detecting}  &  CVPR2022 & EfficientNet-b4  & 99.64   & \textbf{93.18} & 97.40          & 77.70          & 72.42 & 78.92 & 86.54 / 83.92 \\
ICT~\cite{dong2022protecting} &  CVPR2022 & ViT-B  & 90.22    & 85.71          & 95.97          & 93.57          & 76.74 & 77.82 & 86.67 / 85.96 \\
SLADD~\cite{chen2022self} & CVPR2022 & Xception  & 98.40     & 79.70          & 93.73             & 92.80          & 81.80 & 76.94 & 87.23 / 84.99 \\
DCL~\cite{sun2022dual} & AAAI2022 & EfficientNet-b4  & 99.30 & 82.30 & 97.29 & 97.48 & 76.71 & 74.06 & 87.86 / 85.57 \\
CADDM~\cite{Dong_2023_CVPR} & CVPR2023 & ResNet-34  &   99.70  &  91.15 &   97.32   &   81.92   & 71.49 & 81.22 &  87.13 / 84.62 \\
SFDG~\cite{wang2023cvpr_SFDG} & CVPR2023 & ResNet-50  & 95.98      &  75.83         &      \textbf{---}       &      92.10     & 73.64 & \textbf{---}  & \textbf{---}  \\
VLFFD~\cite{sun2023general} & ArXiv2023 & ViT-L & 99.23  & 84.80 & \textbf{---} & \textbf{---} & \textbf{---} & \textbf{---} & \textbf{---} \\
CCDFM~\cite{wang2023exploiting}  & TCSVT2023 & \multicolumn{1}{c}{\textbf{---}}  & 98.50 & 89.10 & \textbf{---} & \textbf{---} & 78.40 & \textbf{---} & \textbf{---} \\
MRL~\cite{yang2023masked} & TIFS2023 & MC3-18  & 98.27 & 83.58 & 92.38 & 88.81 & 71.53 & 76.67 & 85.21 / 82.59 \\
CFM~\cite{Luo2024tifs_CFM} & TIFS2024 & EfficientNet-b4  & 99.62 & 89.65 & 93.34 & 96.38 & 80.22 & 78.06 & 89.55 / 87.53 \\
Zhang24~\cite{zhang2024face} & TCSVT2024 & EfficientNet  & 94.14  & 69.84    & \textbf{---}    & 96.28 & 71.28 & \textbf{---} & \textbf{---} \\
SO-ViT-B~\cite{zou2024semantic} & TIFS2025  & ViT-B & 98.59 & 87.53 & 95.01 & 91.92 & 79.58 & 83.32 & 89.33 / 87.47 \\
\hline
SJEDD (Ours)  & \multicolumn{1}{c}{\textbf{---}} & ViT-B & \textbf{99.86}       & 92.22    & \textbf{97.80} & 93.21   & \textbf{84.14} & \textbf{85.70} & \textbf{92.16} / \textbf{90.61} \\

\bottomrule
\end{tabular}
}
\end{table*}

\section{Experiments}\label{sec: exps}
In this section, we first describe the experimental setups and then compare the proposed SJEDD with contemporary DeepFake detectors. In addition, we perform a series of ablation studies to validate the design choices of SJEDD from both the model and data perspectives.

\subsection{Experimental Setups}\label{subsec:exp_setup}

\subsubsection{Datasets}
We adopt the widely used FF+~\cite{rossler2019faceforensics} and newly established FFSC~\cite{zou2024semantic} for training. FF++ contains $1,000$ real videos,  manipulated by four DeepFake methods: Deepfakes (DF)~\cite{faceswap}, Face2Face (F2F)~\cite{thies2016face2face}, FaceSwap (FS)~\cite{faceswap_Kowalski}, and NeuralTexures (NT)~\cite{thies2019deferred}. These videos come with three compression levels, and our experiments are conducted at the slight compression level, as recommended by~\cite{chen2022self, chen2022ost, haliassos2021lips}.
We expand FF++ using the proposed dataset expansion technique described in Sec.~\ref{subsec:tdg}. In contrast, FFSC~\cite{zou2024semantic} organizes twelve face manipulations into a semantic hierarchical graph with five global face attribute nodes and six local face region nodes, requiring no further dataset expansion. We test DeepFake detectors on four independent datasets: CDF~\cite{li2020celeb}, FaceShifter (FSh)~\cite{Li_2020_CVPR}, DeeperForensics-1.0 (DF-1.0)~\cite{jiang2020deeperforensics}, and DeepFake Detection Challenge (DFDC)~\cite{Dolhansky2020deepfake}.

\subsubsection{Implementation Details}
SJEDD relies on the CLIP model~\cite{radford2021CLIP} for joint embedding, where the image encoder is 
ViT-B/32~\cite{dosovitskiy2020image} and the text encoder is GPT-2~\cite{radford2019GPT-2}. 
SJEDD is fine-tuned by minimizing Problem~\eqref{eq:auto-weighting} using AdamW~\cite{loshchilov2017decoupled} with a decoupled weight decay of $10^{-3}$ and a minibatch size of $32$. The initial learning rate is set to $6\times10^{-7}$, following a cosine annealing schedule~\cite{loshchilov2016sgdr}. The loss weightings are all initialized to one and optimized using Adam~\cite{Kingma2014adam} with a learning rate of $10^{-3}$. Throughout training, the temperature parameter $\tau$ in Eq.~\eqref{eq:unnorm_similarity} is fixed at $10$.
Following \cite{Dong_2023_CVPR, wang2019cnngenerated,zhao2021multi}, we randomly introduce one of five real perturbations to each input image with a probability of $0.3$: patch substitution, additive white Gaussian noise, Gaussian blurring, pixelation, and JPEG compression. These perturbations are constrained to preserve face semantics (\ie, label-preserving). The training process is run for up to $20$ epochs, taking approximately $21$ hours to complete on a single NVIDIA RTX 3090 GPU.

\subsubsection{Competing Methods}~\label{subsubsec: competing_methods}
We compare SJEDD against state-of-the-art DeepFake detectors: CNND~\cite{wang2019cnngenerated}, Face X-ray~\cite{li2020face}, MADD~\cite{zhao2021multi}, Lip-Forensics~\cite{haliassos2021lips}, FRDM~\cite{luo2021generalizing}, RECCE~\cite{Cao_2022_CVPR}, SBI~\cite{shiohara2022detecting}, ICT~\cite{dong2022protecting}, SLADD~\cite{chen2022self}, DCL~\cite{sun2022dual}, CADDM~\cite{Dong_2023_CVPR}, SFDG~\cite{wang2023cvpr_SFDG}, VLFFD~\cite{sun2023general}, CCDFM~\cite{wang2023exploiting}, MRL~\cite{yang2023masked}, CFM~\cite{Luo2024tifs_CFM}, Zhang24~\cite{zhang2024face}, and SO-ViT-B~\cite{zou2024semantic}.
CNND serves as a common baseline. MADD employs attention mechanisms for feature selection and aggregation. FRDM and CCDFM expose face forgery through high-frequency analysis. DCL, CFM, and Zhang24 leverage contrastive learning, while SFDG and MRL utilize graph learning.
Face X-ray, SLADD, and CADDM incorporate manipulation localization as a manipulation-oriented auxiliary task. SLADD and VLFFD additionally include $C$-way manipulation classification. RECCE suggests reconstructing face images to learn generative face features.
Face X-ray, SBI, ICT, SLADD, and VLFFD exploit different data augmentations, including face swapping between real faces~\cite{li2020face, shiohara2022detecting, dong2022protecting} and local face region manipulation~\cite{chen2022self, sun2023general}. 
Lip-Forensics and ICT rely on high-level lipreading and face identity features, respectively. Like the proposed SJEDD, SO-ViT-B employs a semantics-oriented multitask learning paradigm.

\subsubsection{Evaluation Metrics}
In line with~\cite{haliassos2021lips, luo2021generalizing, shiohara2022detecting, dong2022protecting, chen2022self, Dong_2023_CVPR, Cao_2022_CVPR, sun2022dual, zhao2021multi, Yan_2023_ICCV}, we evaluate detection performance using the area under the receiver operating characteristic curve (AUC (\%)).

\subsection{Comparison with Contemporary Detectors}~\label{subsec: exp_compare_detectors}
\subsubsection{Training on FF++}
We train SJEDD on the expanded FF++ dataset~\cite{rossler2019faceforensics} and evaluate it on the test set of the original FF++ as well as five other DeepFake datasets~\cite{li2020celeb, Li_2020_CVPR, jiang2020deeperforensics, Dolhansky2020deepfake, zou2024semantic}.

\noindent\textbf{Cross-dataset evaluation.} Table~\ref{tab:cross_dataset_test} shows the AUC results with several interesting observations. First, many detectors struggle to generalize. In contrast, SJEDD performs well on all five DeepFake datasets, particularly excelling in challenging datasets such as CDF, DFDC, and FFSC. Second, while SBI reports a high AUC of $93.18\%$ on CDF, it performs poorly on DF-1.0 and DFDC.
This performance discrepancy may be due to the heavy reliance on manipulation-specific features, such as statistical inconsistencies in color appearances and landmark locations. Third, all competing methods score relatively low on DFDC, likely resulting from the significant domain shift caused by different imaging conditions and manipulation techniques. 
Last, SJEDD outperforms its predecessor, SO-ViT-B, confirming its improvements in model design (\ie, from discriminative learning to joint embedding) and loss function (\ie, from the cross-entropy to fidelity loss in bi-level optimization).

\begin{table}[!t]
\renewcommand{\arraystretch}{1.2}
\caption{AUC results in the cross-manipulation setting. 
DF, F2F, FS, and NT represent Deepfakes~\cite{faceswap}, Face2Face~\cite{thies2016face2face}, FaceSwap~\cite{faceswap_Kowalski}, and NeuralTexures~\cite{thies2019deferred}, respectively. Gray numbers represent within-manipulation results}
\label{tab:cross_manipulation_test}
\small
\centering
\resizebox{\linewidth}{!}{
\begin{tabular}{clccccc}
\toprule
Training        & Method     & DF & F2F & FS & NT & Mean AUC \\
\hline
\multirow{7}{*}{DF}  & MADD~\cite{zhao2021multi}  &  \textcolor{gray}{99.51}  &  66.41   & 67.33   &  66.01  &  66.58    \\
                     & RECCE~\cite{Cao_2022_CVPR} & \textcolor{gray}{99.65}   &  70.66   &  74.29  &  67.34  & 70.76     \\
                     & DCL~\cite{sun2022dual}     &  \textcolor{gray}{99.98}  &  77.13   &  61.01  &  75.01  & 71.05     \\
                     & VLFFD~\cite{sun2023general} &  \textcolor{gray}{99.97}  &  87.46   &  74.40  &  76.79  & 79.55    \\
                     & CFM~\cite{Luo2024tifs_CFM} & \textcolor{gray}{99.93} & 77.56 & 54.94 & 75.04 & 69.18 \\
                     & SO-ViT-B~\cite{zou2024semantic} &  \textcolor{gray}{99.39}  &  \textbf{88.25} &  89.31  &  82.67  & 86.74 \\
                     \cline{2-7}
                     & SJEDD (Ours)      &  \textcolor{gray}{99.51}  & 83.89  & \textbf{95.02} &  \textbf{83.28}  &   \textbf{87.40}  \\
\hline
\multirow{7}{*}{F2F} & MADD~\cite{zhao2021multi}       &  73.04  &   \textcolor{gray}{97.96}  &  65.10  &  71.88  & 70.01 \\
                     & RECCE~\cite{Cao_2022_CVPR}      & 75.99   &  \textcolor{gray}{98.06}   &  64.53  &  72.32  & 70.95  \\
                     & DCL~\cite{sun2022dual}       &  91.91  &  \textcolor{gray}{99.21}   &  59.58  &  66.67  & 72.72  \\
                     & VLFFD~\cite{sun2023general} &  94.90  &  \textcolor{gray}{99.30}   &  65.19  & 66.69   &  75.59 \\
                     & CFM~\cite{Luo2024tifs_CFM} & 81.85 & \textcolor{gray}{99.23} & 60.12 & 70.80 & 70.92 \\
                     & SO-ViT-B~\cite{zou2024semantic} &  99.68  &  \textcolor{gray}{99.39}   &  \textbf{97.70}  & 92.05   &  96.47 \\
                     \cline{2-7}
                     & SJEDD (Ours) & \textbf{99.83} &  \textcolor{gray}{99.46}   &  97.45  & \textbf{92.34}  &  \textbf{96.54}    \\
\hline
\multirow{7}{*}{FS}  & MADD~\cite{zhao2021multi}       &  82.33  &  61.65   &  \textcolor{gray}{98.82}  &  54.79  &  66.26    \\
                     & RECCE~\cite{Cao_2022_CVPR}      &  82.39  &  64.44   &  \textcolor{gray}{98.82}  &  56.70  &  67.84    \\
                     & DCL~\cite{sun2022dual}       &  74.80  &  69.75   &  \textcolor{gray}{99.90}  &  52.60  &  65.72    \\
                     & VLFFD~\cite{sun2023general} &  92.98  &  79.69   & \textcolor{gray}{99.57}   &  53.53  & 75.40     \\
                     & CFM~\cite{Luo2024tifs_CFM} & 72.91 & 71.39 & \textcolor{gray}{99.85} & 51.69 & 65.33 \\
                     & SO-ViT-B~\cite{zou2024semantic} &  98.88  &  76.48   & \textcolor{gray}{96.07}   &  75.38  & 83.58     \\
                     \cline{2-7}
                     & SJEDD (Ours) &  \textbf{99.68}  &  \textbf{81.84}   &  \textcolor{gray}{93.29}  &  \textbf{81.19}  &   \textbf{87.57}   \\
\hline
\multirow{7}{*}{NT}  & MADD~\cite{zhao2021multi} &  74.56  &  80.61   &  60.90  & \textcolor{gray}{93.34}   &  72.02    \\
                     & RECCE~\cite{Cao_2022_CVPR} &  78.83  &  80.89   &  63.70  &  \textcolor{gray}{93.63}  & 74.47     \\
                     & DCL~\cite{sun2022dual}    &  91.23  &  52.13   &  79.31  &  \textcolor{gray}{98.97}  &  74.22    \\
                     & VLFFD~\cite{sun2023general} &  92.74  &  62.53   &  85.62  &  \textcolor{gray}{98.99}  &  80.30    \\
                     & CFM~\cite{Luo2024tifs_CFM} & 88.31 & 76.78 & 52.56 & \textcolor{gray}{99.24} & 72.55 \\
                     & SO-ViT-B~\cite{zou2024semantic} &  99.40  &  94.86   &  94.25  &  \textcolor{gray}{95.11}  &  96.16    \\
                     \cline{2-7}
                     & SJEDD (Ours) & \textbf{99.88}  &  \textbf{95.14} &  \textbf{97.31}  &  \textcolor{gray}{98.51}  &  \textbf{97.44} \\
\bottomrule
\end{tabular}}
\end{table}

\noindent \textbf{Cross-manipulation evaluation.} We perform the cross-manipulation evaluation of SJEDD against six representative DeepFake detectors: MADD~\cite{zhao2021multi}, RECCE~\cite{Cao_2022_CVPR}, DCL~\cite{sun2022dual}, VLFFD~\cite{sun2023general}, CFM~\cite{Luo2024tifs_CFM}, and SO-ViT-B~\cite{zou2024semantic}.
Specifically, we train on one manipulation type from FF++~\cite{rossler2019faceforensics} (including additional augmented data) and test on the remaining manipulations from FF++. 
As shown in Table~\ref{tab:cross_manipulation_test}, SJEDD surpasses other methods by clear margins, which we believe arises from two primary reasons. First, our semantics-oriented formulation helps SJEDD learn generalizable features across different manipulations that alter the same face attributes (\eg, NT $\rightarrow$ F2F, both modifying the global \texttt{expression} attribute and the local \texttt{mouth} and \texttt{lip} regions). Second, SJEDD models the joint probability distribution of face attributes (see Eq.~\eqref{eq:plf_joint}), thus promoting generalization across different face attributes (\eg, DF: \texttt{identity} $\rightarrow$ NT: \texttt{expression}). Once again, SJEDD outperforms its predecessor, SO-ViT-B, highlighting the effectiveness of the joint embedding formulation and the fidelity loss in bi-level optimization.

\noindent\textbf{Cross-attribute evaluation.}
As suggested in FFSC~\cite{zou2024semantic}, we adopt two semantics-oriented testing protocols: Protocol-1, which focuses on generalization to novel manipulations for the same face attribute; and Protocol-2, which examines generalization to novel face attributes. Six DeepFake detectors are chosen for comparison, including Lip-Forensics~\cite{haliassos2021lips}, SLADD~\cite{chen2022self}, ICT~\cite{dong2022protecting}, DCL~\cite{sun2022dual}, CFM~\cite{Luo2024tifs_CFM}, and SO-ViT-B~\cite{zou2024semantic}. 
The results on the FFSC main test set are shown in Table~\ref{tab: semantics_test}, where we find that SLADD, DCL, and CFM perform poorly under Protocol-1. This provides a strong indication that the features they learn are manipulation-specific and thus ineffective at detecting the same face attribute forgery by different methods. In stark contrast, semantics-oriented methods such as SO-ViT-B and the proposed SJEDD yield favorable results in the Protocol-1 test, underscoring the advantages of learning and integrating semantic features at the global face attribute level with signal features at the local face region level.
When switching to Protocol-2, SJEDD continues to outperform manipulation-oriented methods, particularly in generalizing to the unseen \texttt{gender} attribute. 

\begin{table*}[!t]
\renewcommand{\arraystretch}{1.1}
\caption{AUC results in the cross-attribute setting using the FFSC main test set. All models are trained on the (respectively augmented) training set of FF++}
\label{tab: semantics_test}
\small
\centering
\resizebox{0.78\linewidth}{!}{
\begin{tabular}{lcccccccc}
\toprule
\multirow{2}{*}{Method} & \multirow{2}{*}{Intra-dataset} & \multicolumn{2}{c}{Protocol-1} & \multicolumn{3}{c}{Protocol-2} & \multirow{2}{*}{Mean AUC}   \\
\cmidrule(lr){3-4} \cmidrule(lr){5-7}
                    &    & \texttt{Expression}      & \texttt{Identity}     & \texttt{Age}      & \texttt{Gender}    & \texttt{Pose}   &                     \\
\hline
Lip-Forensics~\cite{haliassos2021lips} & \textcolor{gray}{99.70} & \textbf{---} & 76.76 & 60.75 & 76.76 & \textbf{---} & \textbf{---} \\              
SLADD~\cite{chen2022self} & \textcolor{gray}{98.40} & \textbf{82.39} & 65.55 & 85.31 & 81.21 & 70.00 & 76.89 \\
ICT~\cite{dong2022protecting} & \textcolor{gray}{90.22} & \textbf{---} & 84.20 & 72.18 & 52.83 & 76.12 & \textbf{---} \\
DCL~\cite{sun2022dual} & \textcolor{gray}{99.30} & 65.72 & 85.25 & 84.11 & 75.41 & 71.67 & 76.43 \\
CFM~\cite{Luo2024tifs_CFM} & \textcolor{gray}{99.62} & 73.37 & 89.97 & \textbf{86.76} & 77.68 & 74.59 & 80.47 \\
SO-ViT-B~\cite{zou2024semantic} & \textcolor{gray}{98.59} & 76.89 & \textbf{92.83} & 86.39 & 81.25 & 88.42 & 85.16 \\
\hline
SJEDD (Ours)    & \textcolor{gray}{99.86} & 80.16 & 92.23 & 86.45 & \textbf{86.37} & \textbf{89.03} & \textbf{86.85} \\
\bottomrule
\end{tabular}
}
\end{table*}

\begin{table}[!t]
\renewcommand{\arraystretch}{1.1}
\caption{AUC results of SJEDD against six contemporary DeepFake detectors in the cross-dataset setting. All models are trained on FFSC
}
\label{tab: FFSC_train_test}
\centering
\begin{tabular}{lccccc}
\toprule
Method       & FF++ & CDF & DF-1.0 & DFDC & Mean AUC \\
\hline
CNND~\cite{wang2019cnngenerated}      & 80.57 & 83.93 & 85.98 & 72.99 & 80.87 \\

MADD~\cite{zhao2021multi}      & 87.63 & 87.46 & 83.96 & 77.02 & 84.02 \\

FRDM~\cite{luo2021generalizing}      & 89.07 & 80.39 & 88.46 & 73.58 & 82.87  \\

RECCE~\cite{Cao_2022_CVPR}     & \textbf{95.56} & 78.51 & 69.73 & 67.20 & 77.77 \\

CADDM~\cite{Dong_2023_CVPR}     & 83.29 & 81.10 & 87.04 & 65.23 & 79.17 \\

SO-ViT-B~\cite{zou2024semantic}  & 86.33 & 88.43 & 92.46 & 79.57 & 86.77 \\
\hline
SJEDD (Ours)         & 93.81 & \textbf{91.28} & \textbf{97.60} & \textbf{81.32} & \textbf{91.00} \\
\bottomrule
\end{tabular}
\end{table}

\subsubsection{Training on FFSC}
We next train the proposed SJEDD on FFSC~\cite{zou2024semantic}, which encompasses five global face attributes---\texttt{age}, \texttt{expression}, \texttt{gender}, \texttt{identity}, and  \texttt{pose}---instantiated by twelve face manipulations. The connections between global face attribute nodes and local face region nodes, as well as the manipulation degree parameters, are determined through formal psychophysical experiments~\cite{zou2024semantic}. The text templates to embed the label hierarchy are identical to those described in Sec.~\ref{subsec: joint_embedding}.
We conduct the cross-dataset evaluation of SJEDD against six DeepFake detectors: CNND~\cite{wang2019cnngenerated}, MADD~\cite{zhao2021multi}, FRDM~\cite{luo2021generalizing}, RECCE~\cite{Cao_2022_CVPR}, CADDM~\cite{Dong_2023_CVPR}, and SO-ViT-B~\cite{zou2024semantic} on FF++~\cite{rossler2019faceforensics}, CDF~\cite{li2020celeb}, DF-1.0~\cite{jiang2020deeperforensics}, and DFDC~\cite{Dolhansky2020deepfake}. All competing methods are trained on FFSC.

As shown in Table~\ref{tab: FFSC_train_test}, SJEDD performs much better than the competing detectors on the four unseen DeepFake datasets. This shows the effectiveness and generality of the proposed semantics-oriented joint embedding in enhancing DeepFake detection. 
Additionally, all detectors demonstrate strong performance on FF++, with RECCE achieving an impressive AUC of $95.56\%$. This is due to the relatively small domain gap between FF++ and FFSC, as a portion of images in these datasets share the same source (\ie, YouTube).  Consequently, RECCE may easily ``shortcut'' through face reconstruction. In contrast, the detection accuracy on DFDC is comparatively lower, consistent with the observations in Table~\ref{tab:cross_dataset_test}.

\subsection{Ablation Studies}
In this subsection, we conduct a series of ablation experiments to verify SJEDD's design choices from both the model and data perspectives.

\subsubsection{Ablation from the Model Perspective}~\label{subsec: abaltions_model}
We first analyze two key components of SJEDD:  the joint embedding in Eq.~\eqref{eq:unnorm_similarity} and the fidelity loss in Eq.~\eqref{eq: loss_b}. Specifically, we construct two SJEDD degenerates: one implemented solely by the image encoder to produce the raw scores (\ie, $\bm s(\bm x) = \bm f_{\bm \phi} (\bm x)$) and trained with the cross-entropy loss, and the other implemented by joint embedding and trained with the cross-entropy loss. Table~\ref{tab: sjedd_model} shows the cross-dataset evaluation results, in which all model variants are trained on FFSC.

\begin{figure}[!t]
  \centering
  \subfloat[Pretrained CLIP]{\includegraphics[width=1\linewidth]{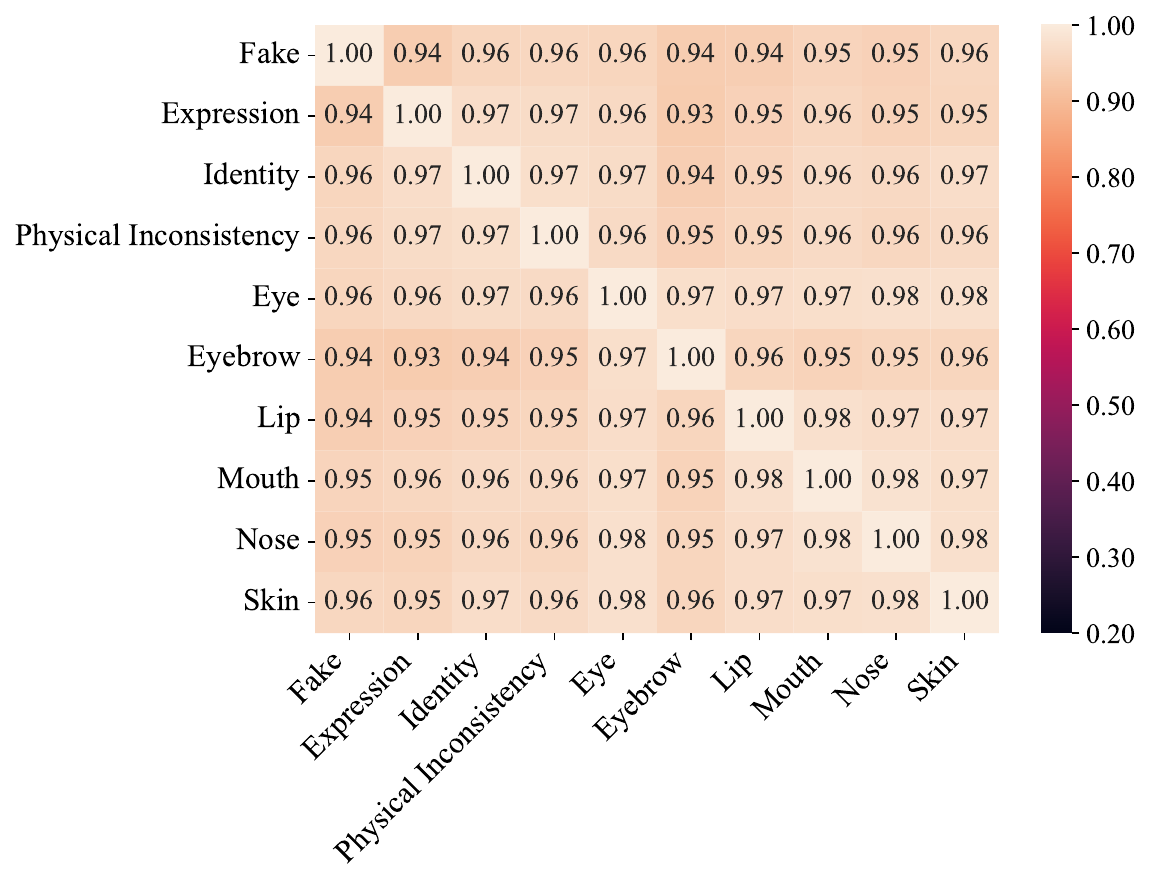}} 
  \\
  \subfloat[SJEDD]{\includegraphics[width=1\linewidth]{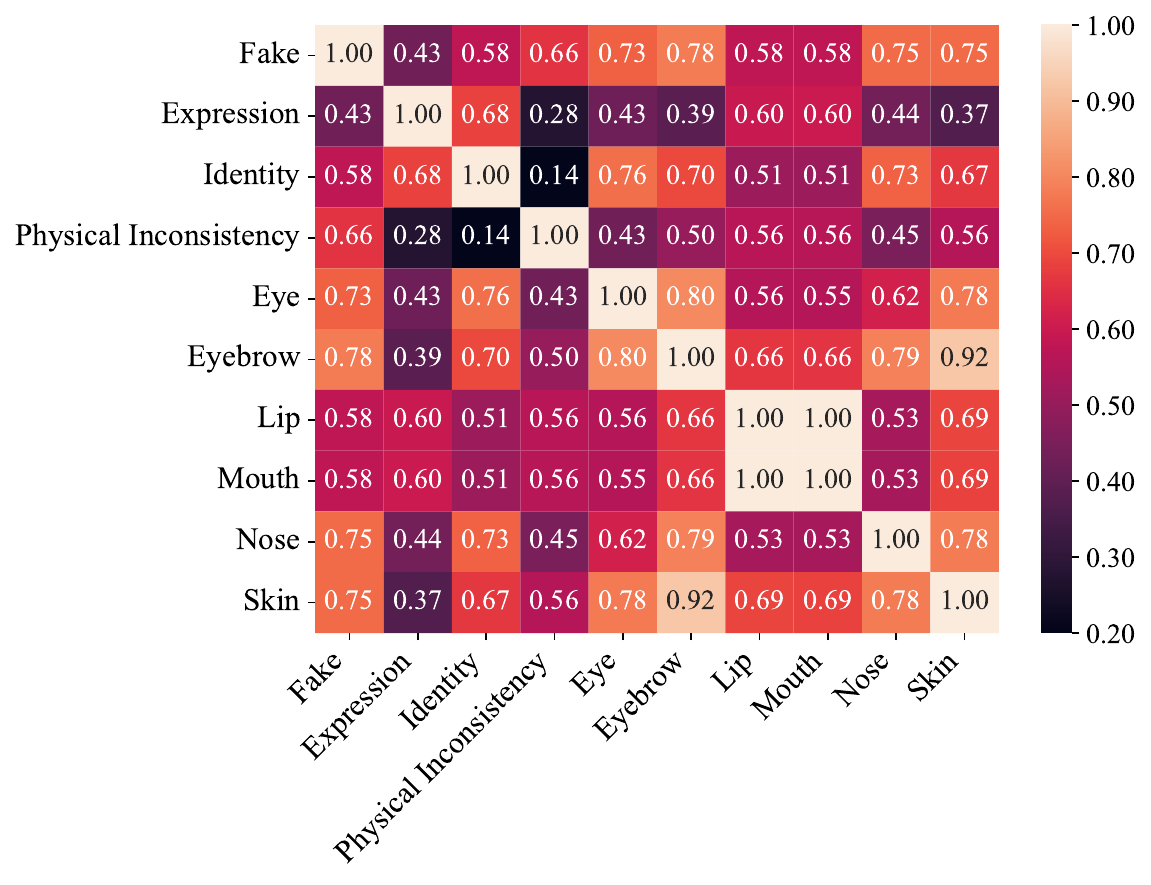}}
  \caption{
  Illustration of semantic relationships among tasks before and after semantics-oriented multitask learning. The relationships are shown using a correlation matrix, in which each entry represents the cosine similarity between two task-specific textual embeddings. Zoom in for improved visibility.
  }
  \label{fig: matrix_task_relationships}
\end{figure}

\begin{table}[!t]
\renewcommand{\arraystretch}{1.2}
\setlength{\abovecaptionskip}{0cm}
\caption{AUC results of two degenerates of SJEDD. The term ``w/o joint embedding'' refers to directly predicting the label hierarchy using only the image encoder. ``w/o fidelity loss'' means substituting the fidelity loss with the cross-entropy loss during training. All models are trained on FFSC}
\label{tab: sjedd_model}
\centering
\resizebox{\linewidth}{!}{
\begin{tabular}{ccccccc}
\toprule
Joint Embedding & Fidelity Loss & FF++ & CDF & DF-1.0 & DFDC & Mean AUC  \\
\hline
\noalign{\smallskip}
\xmark & \xmark & 86.63 & 88.43 & 92.46 & 79.57 & 86.77 \\
\cmark & \xmark  & 91.77 & 90.23 & 96.22 & \textbf{81.35} & 89.89\\
\hline
\cmark & \cmark  & \textbf{93.81} & \textbf{91.28} & \textbf{97.60} & 81.32 & \textbf{91.00} \\
\bottomrule
\end{tabular}
}
\end{table}

\begin{figure}[!t]
  \centering
  \includegraphics[width=0.35\textwidth]{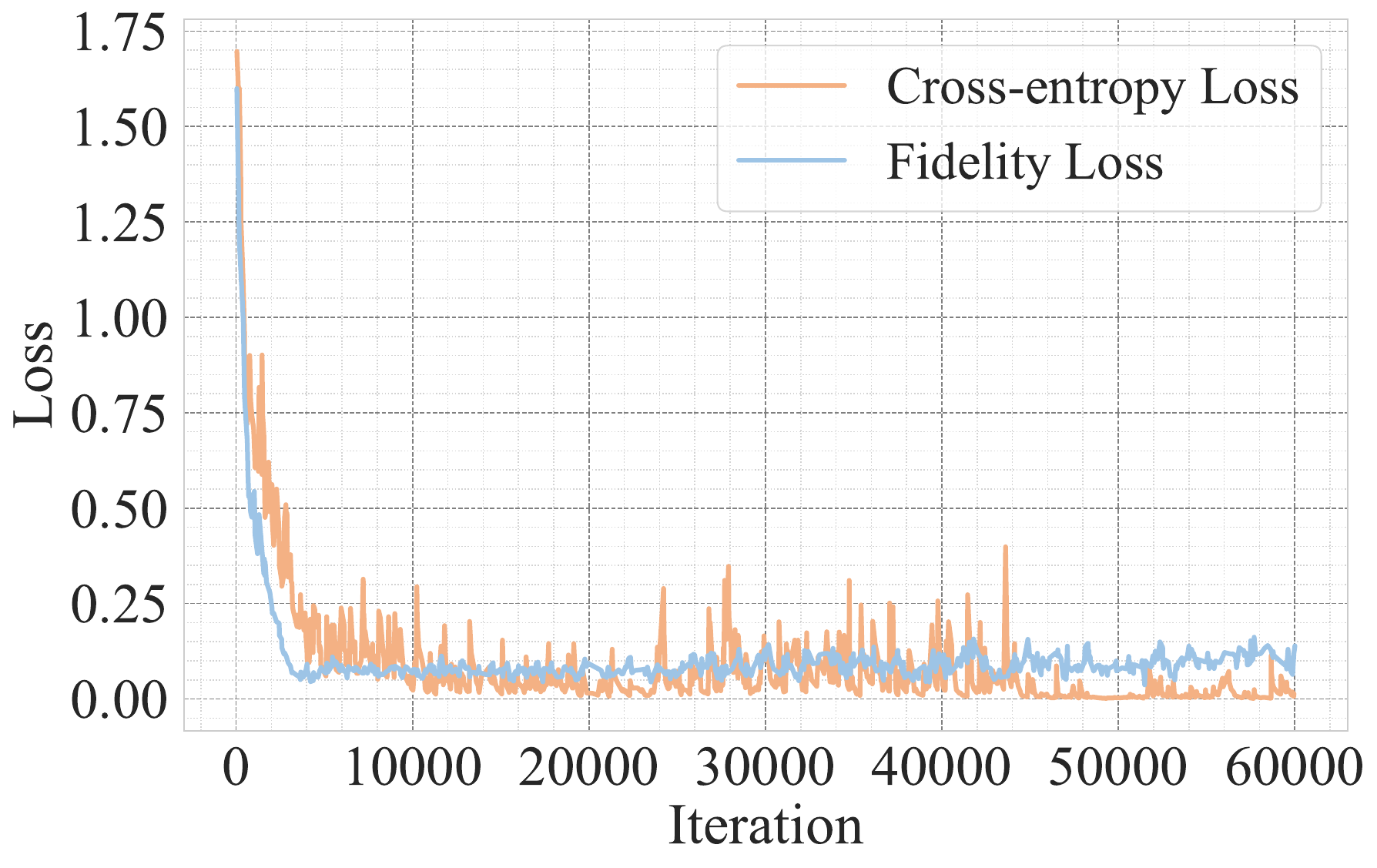}
  \caption{Training dynamics induced by the cross-entropy and fidelity losses.}
  \label{fig: loss-dynamic}
\end{figure}

It is clear from the table that joint embedding achieves noticeably better performance than directly predicting the label hierarchy from the input face image. This highlights the importance of adjusting model capacity for each task through end-to-end optimization. Additionally, joint embedding facilitates task relationship modeling. 
As shown in Fig.~\ref{fig: matrix_task_relationships}, initially, the pretrained CLIP text encoder does not adequately capture semantic similarities among tasks, treating most tasks equally. However, after optimizing for joint embedding, the relationships between tasks are effectively learned. For example, the concept of ``expression'' becomes more closely associated with ``lip'' and ``mouth.'' 
By further incorporating the fidelity loss, SJEDD achieves the best results. 
Fig.~\ref{fig: loss-dynamic} illustrates the training dynamics of the two loss functions, where we find that the cross-entropy loss exhibits pronounced fluctuations, whereas the fidelity loss curve is more stable. 

\begin{figure*}[t]
  \centering
  \subfloat[FF++: Real]{\includegraphics[width=0.32\linewidth]{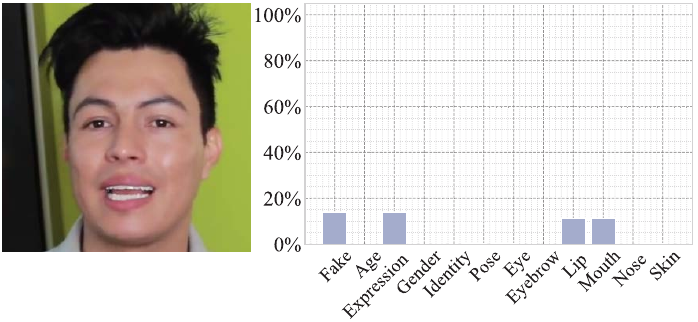}} \hskip.2em
  \subfloat[CDF: Real]{\includegraphics[width=0.32\linewidth]{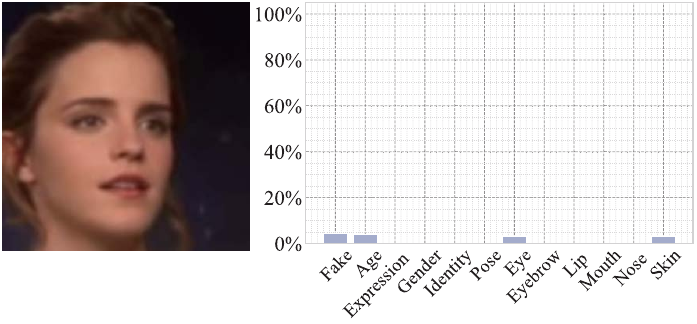}} \hskip.2em
  \subfloat[DFDC: Real]{\includegraphics[width=0.32\linewidth]{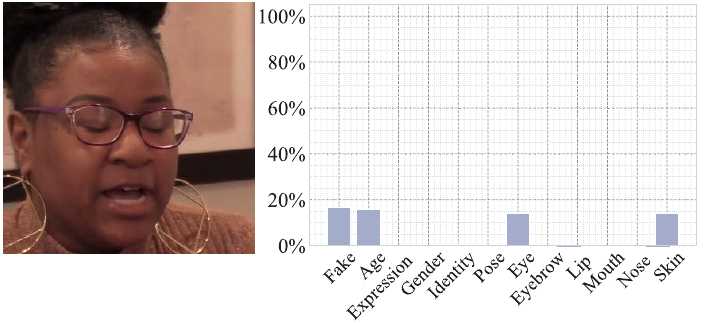}}
  \\
  \subfloat[FFSC: \texttt{Age}]{\includegraphics[width=0.32\linewidth]{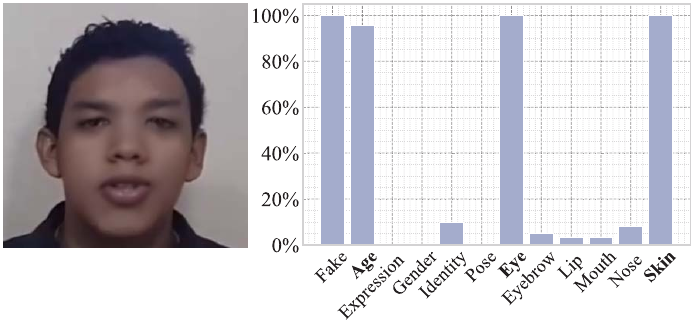}} \hskip.2em
  \subfloat[FFSC: \texttt{Expression} (Smile)]{\includegraphics[width=0.32\linewidth]{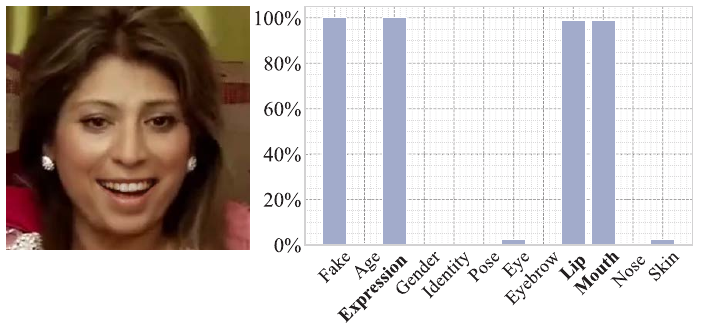}} \hskip.2em
  \subfloat[FFSC: \texttt{Expression} (Surprise)]{\includegraphics[width=0.32\linewidth]{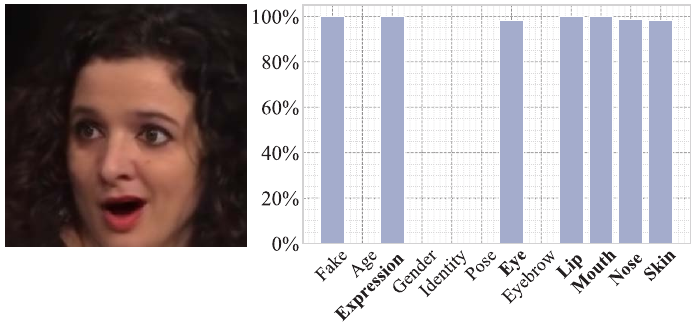}} \hskip.2em
  \\
  \subfloat[FFSC: \texttt{Gender}]{\includegraphics[width=0.32\linewidth]{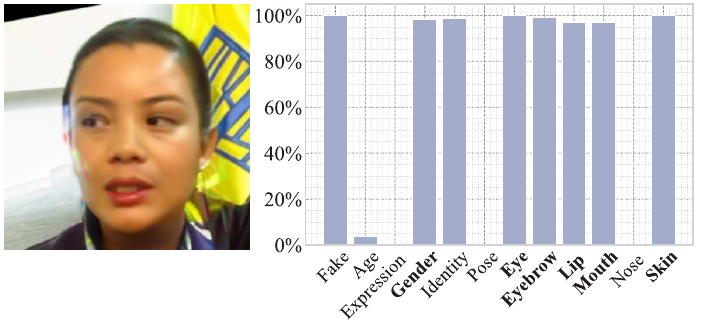}} \hskip.2em
  \subfloat[FFSC: \texttt{Identity}]{\includegraphics[width=0.32\linewidth]{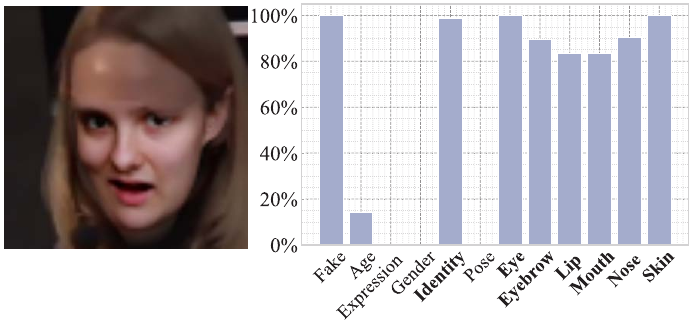}} \hskip.2em
  \subfloat[FFSC: \texttt{Pose}]{\includegraphics[width=0.32\linewidth]{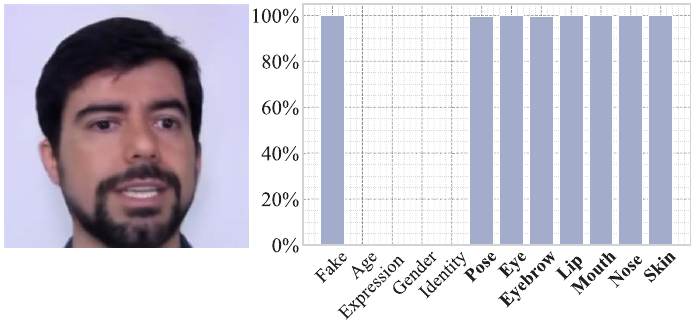}}
  \\
  \subfloat[FF++: DF]{\includegraphics[width=0.32\linewidth]{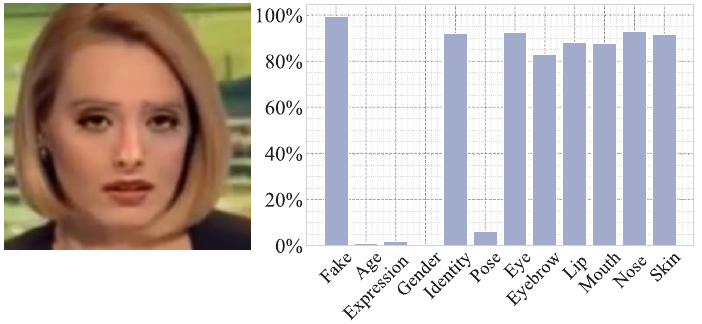}} \hskip.2em
  \subfloat[FF++: F2F]{\includegraphics[width=0.32\linewidth]{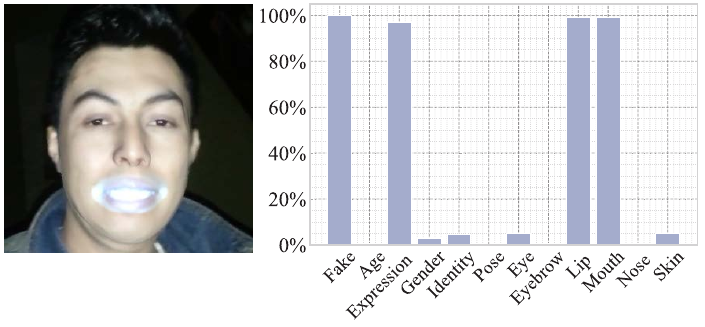}} \hskip.2em
  \subfloat[FF++: FS]{\includegraphics[width=0.32\linewidth]{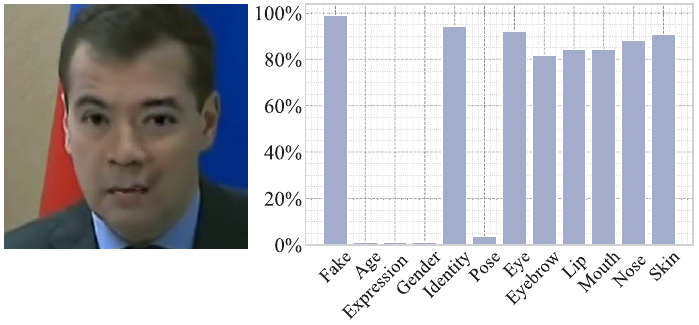}}
  \\
  \subfloat[FF++: NT]{\includegraphics[width=0.32\linewidth]{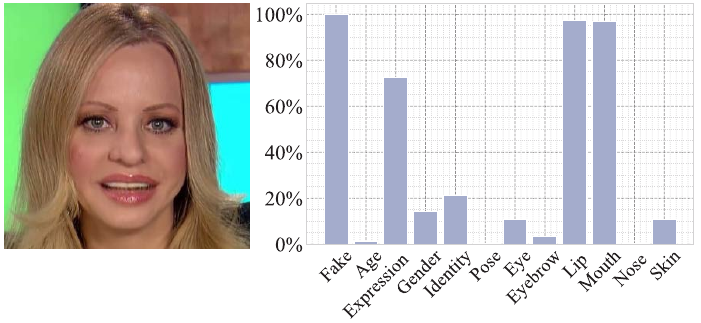}} \hskip.2em
  \subfloat[CDF]{\includegraphics[width=0.32\linewidth]{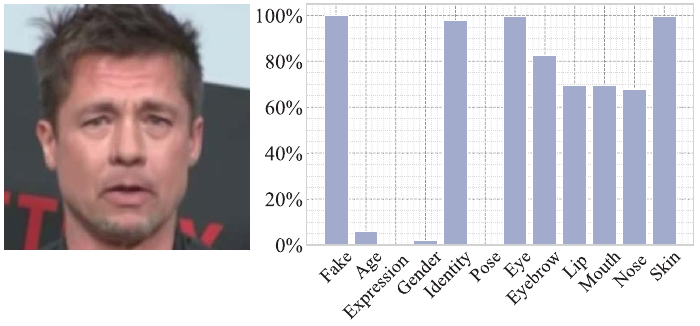}} \hskip.2em
  \subfloat[DFDC]{\includegraphics[width=0.32\linewidth]{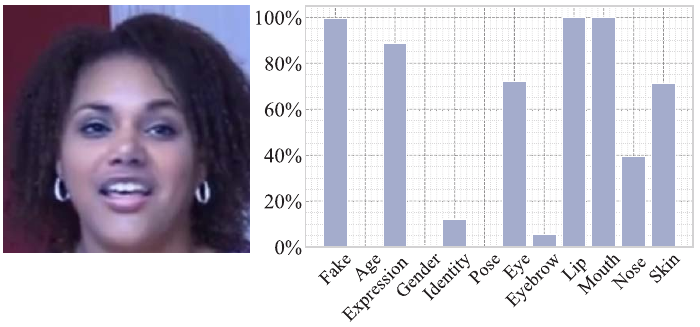}}

  \caption{Illustration of the proposed SJEDD trained on FFSC in making predictions at the face attribute and region levels. 
  For FFSC (as an intra-dataset test), fake attributes/regions are highlighted in bold.
  For FF++, CDF, and DFDC (as cross-dataset tests),  only prediction results are provided for reference.
  Zoom in for improved visibility.
  }
  \label{fig: visual_language_corres}

\end{figure*}

We then ablate other design choices of SJEDD by 1) freezing the text encoder $\bm g_{\bm \varphi}$, 2) simplifying the text templates in Sec.~\ref{subsec: joint_embedding} to keywords, \ie, ``\textit{fake},'' ``\textit{altered $\{$g$\}$},'' and ``\textit{altered $\{$l$\}$},'' 3) estimating the raw scores in Eq.~\eqref{eq:plf_joint} by computing the cosine (\ie, normalized) similarity between the visual and textual embeddings:
\begin{align}
    s_i(\bm x)= \frac{\langle\bm f_{\bm \phi}(\bm x),\bm g_{\bm \varphi}(\bm t_i)\rangle}{\tau \Vert \bm f_{\bm \phi}(\bm x) \Vert_2 \Vert\bm g_{\bm \varphi}(\bm t_i) \Vert_2},
\end{align}
where $\Vert\cdot\Vert_2$ denotes the $\ell_2$-norm and $\tau$ is the same temperature parameter as in Eq.~\eqref{eq:unnorm_similarity}, 4) optimizing SJEDD with a fixed set of loss weightings $\{1.0, 0.1,0.1\}$ in Eq.~\eqref{eq:overallloss} to prioritize the primary task manually~\cite{Cao_2022_CVPR, chen2022self, sun2022dual, Dong_2023_CVPR, yang2023masked}, 5) optimizing SJEDD with fixed equal loss weightings $\{1.0, 1.0,1.0\}$ in Eq.~\eqref{eq:overallloss}, and 6) training SJEDD with dynamic weighting average~\cite{liu2019end} without prioritizing the primary task.
\begin{table}[!t]
\renewcommand{\arraystretch}{1.2}
\setlength{\tabcolsep}{2pt}
\setlength{\abovecaptionskip}{0cm}
\caption{AUC results of different variants of SJEDD, all trained on the expanded FF++}
\label{tab: ablations}
\centering
\resizebox{\linewidth}{!}{
\begin{tabular}{lccccc}
\toprule
Model Variant      & CDF & DF-1.0 & DFDC & FFSC & Mean AUC  \\
\hline
\noalign{\smallskip}
w/ Frozen $\bm g_{\bm\varphi}$ & \textbf{92.38} & 91.98 & 81.54 & 83.08 & 87.25 \\
w/ Simplified Text Templates & 90.49 & 92.15 & 83.83 & 85.33 & 87.95 \\
w/ Normalized Similarity  & 89.00 & 93.15 & 84.01 & 83.96 & 87.53 \\
w/ Loss Weightings of \{1.0, 0.1, 0.1\}  & 88.90 & 92.04 & 79.94 & 83.75 & 86.16 \\
w/ Loss Weightings of \{1.0, 1.0, 1.0\} & 88.32 & 92.93 & 82.27 & 81.48 & 86.25 \\
w/ Dynamic Weighting Average~\cite{liu2019end}  & 89.02 & \textbf{93.38} & 82.06 & 82.16 & 86.66 \\
\hline
SJEDD (Ours)  &  92.22 & 93.21 & \textbf{84.14} & \textbf{85.70} & \textbf{88.82} \\
\bottomrule
\end{tabular}
}
\end{table}
Table~\ref{tab: ablations} presents the cross-dataset results. First, freezing both the image and text encoders proves ineffective. Fine-tuning the image encoder significantly boosts detection performance, which validates the importance of vision-language alignment. Joint fine-tuning of the image and text encoders offers additional performance gains.
Besides, SJEDD is fairly robust to changes in text template. Using normalized similarity for raw score calculation negatively impacts detection performance, indicating that some discriminative information may be captured in visual embedding magnitudes. Finally, dynamically adjusting the loss weightings and prioritizing the primary task through bi-level optimization proves advantageous, which also removes the need for cumbersome hyperparameter tuning.

\begin{figure*}[!t]
  \centering
  \subfloat[FF++: Real]{\includegraphics[width=0.25\linewidth]{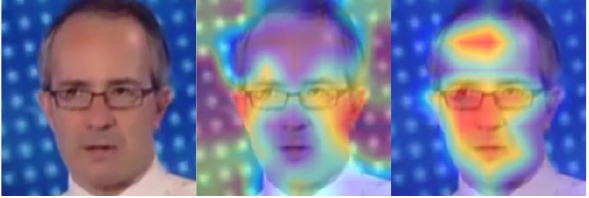}} 
  \subfloat[CDF: Real]{\includegraphics[width=0.25\linewidth]{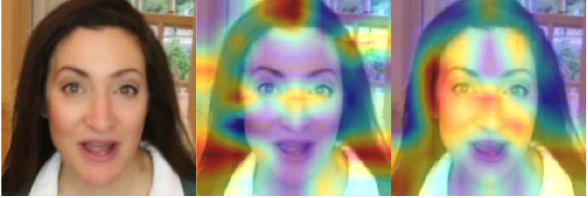}} 
  \subfloat[DFDC: Real]{\includegraphics[width=0.25\linewidth]{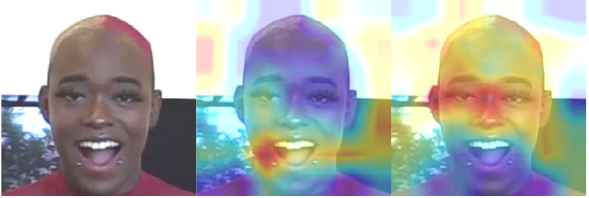}} 
  \subfloat[FFSC: Real]{\includegraphics[width=0.25\linewidth]{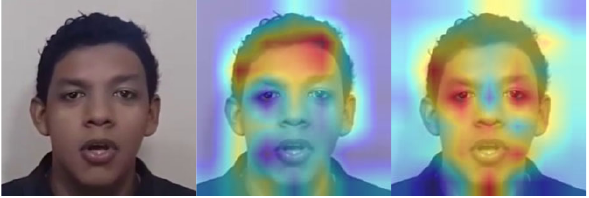}}
  \\
  \subfloat[FF++: DF]{\includegraphics[width=0.25\linewidth]{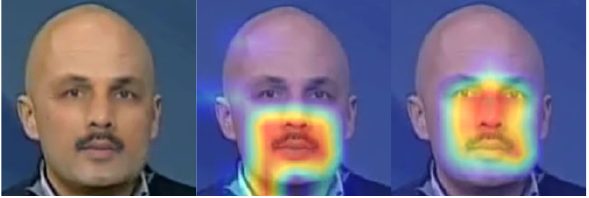}}
  \subfloat[FF++: F2F]{\includegraphics[width=0.25\linewidth]{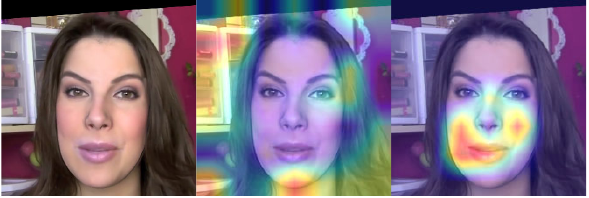}}
  \subfloat[FF++: FS]{\includegraphics[width=0.25\linewidth]{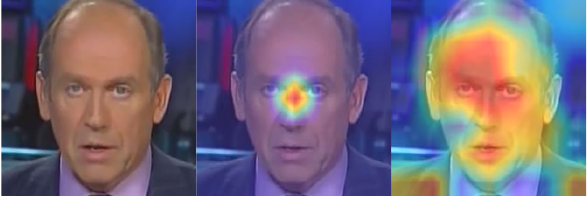}} 
  \subfloat[FF++: NT]{\includegraphics[width=0.25\linewidth]{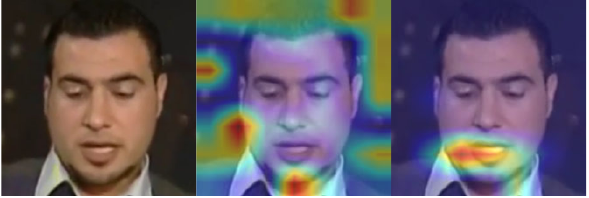}} 
  \\
  \subfloat[CDF]{\includegraphics[width=0.25\linewidth]{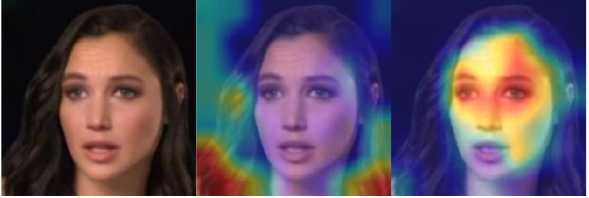}} 
  \subfloat[FSh]{\includegraphics[width=0.25\linewidth]{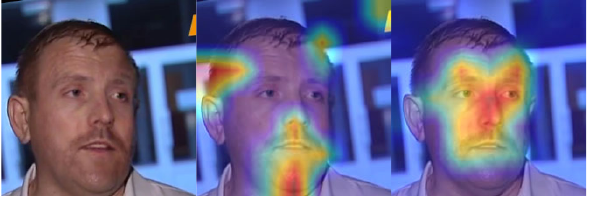}} 
  \subfloat[DF-1.0]{\includegraphics[width=0.25\linewidth]{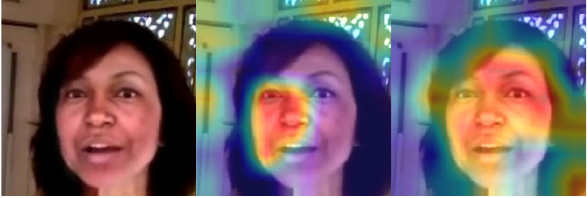}} 
  \subfloat[DFDC]{\includegraphics[width=0.25\linewidth]{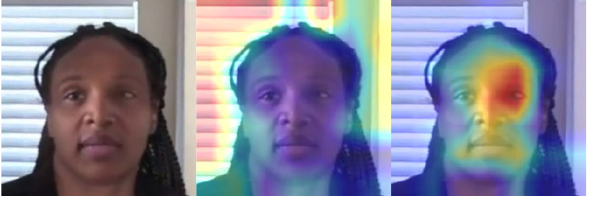}}
  \caption{
  Visualization of Grad-CAM activation maps. Each subfigure shows a forged face (left), baseline heatmap (middle), and SJEDD heatmap (right). Best viewed in color.}
  \label{fig: attention_heatmap}
\end{figure*}

Since SJEDD is built upon CLIP, we also compare it against recent CLIP-based DeepFake detectors, including CLIP~\cite{radford2021CLIP}, VLFFD~\cite{sun2023general}, GM-DF~\cite{lai2024gmdf}, FCG~\cite{han2024fcg}, CLIPPing~\cite{khan2024clipping}, and RepDFD~\cite{lin2024Repdfd}. Table~\ref{tab: clip_based_methods} presents the cross-dataset results, where SJEDD consistently outperforms these CLIP-based detectors, further demonstrating the non-trivial adaptation of CLIP in SJEDD for DeepFake detection.

\begin{table}[!t]
\centering
\caption{
AUC results of the proposed SJEDD against recent CLIP-based DeepFake detectors in the cross-dataset setting. All results are copied directly from the original papers except for the zero-shot CLIP
}
\label{tab: clip_based_methods}
\small
\resizebox{\linewidth}{!}{
\begin{tabular}{llcccc}
\toprule
Method & Venue & Backbone & CDF & DFDC & Mean AUC\\
\midrule
CLIP~\cite{radford2021CLIP}& ICML2021 & ViT-L & 77.70 & 74.20 & 75.95 \\
VLFFD~\cite{sun2023general} & ArXiv2023 & ViT-L & 84.80 & \textbf{---} & \textbf{---} \\
GM-DF~\cite{lai2024gmdf} & ArXiv2024 & ViT-B & 83.20 & 77.20 & 80.20 \\
FCG-L~\cite{han2024fcg} & ArXiv2024 & ViT-L & \textbf{92.30} & 81.20 & 86.75 \\
FCG-B~\cite{han2024fcg} & ArXiv2024 & ViT-B & 83.20 & 75.60 & 79.40 \\
CLIPPing~\cite{khan2024clipping} & ICMR2024 & ViT-L & \textbf{---} & 71.90 & \textbf{---} \\
RepDFD~\cite{lin2024Repdfd} & AAAI2025 & ViT-L & 89.90 & 81.00 & 85.45 \\
\hline
SJEDD (Ours) & \multicolumn{1}{c}{\textbf{---}} & ViT-B & 92.22 & \textbf{84.14} & \textbf{88.18} \\
\bottomrule
\end{tabular}
}
\end{table}

Furthermore, though SJEDD is trained on a GAN-based dataset (\ie, FF++~\cite{rossler2019faceforensics}), we assess its ability to detect diffusion-based face forgery using DiffusionFace~\cite{chen2024diffusionface} and DiFF~\cite{cheng2024diffusion}. As shown in Table~\ref{tab: performance_diffusion_data}, SJEDD achieves strong performance and surpasses other state-of-the-art detectors. These results further validate the effectiveness of our semantics-oriented approach in learning more generalizable features beyond manipulation-based artifacts~\cite{luo2021generalizing, Dong_2023_CVPR}.

\begin{table}[!t]
\centering
\caption{
AUC results on the two diffusion-based DeepFake datasets. ``$\dagger$'' indicates that we exclude purely synthesized faces in these datasets, as they are beyond the scope of our current study}
\label{tab: performance_diffusion_data}
\small
\resizebox{0.95\linewidth}{!}{
\begin{tabular}{lccc}
\toprule
Method & DiffusionFace$^\dagger$~\cite{chen2024diffusionface} & DiFF$^\dagger$~\cite{cheng2024diffusion} & Mean AUC \\
\midrule
MADD~\cite{zhao2021multi} & 74.72 & \textbf{87.35} & 81.04  \\
FRDM~\cite{luo2021generalizing}  & 73.23 & 55.22 &  64.23  \\
RECCE~\cite{Cao_2022_CVPR} & 78.64 & 82.00 &  80.32  \\
CADDM~\cite{Dong_2023_CVPR} & 62.94 & 60.20 &  61.57  \\
SO-ViT-B~\cite{zou2024semantic} & 90.55 & 83.59 & 87.07 \\
\hline
SJEDD (Ours) & \textbf{93.70} & 84.64 & \textbf{89.17} \\
\bottomrule
\end{tabular}
}
\end{table}

\subsubsection{Ablation from the Data Perspective}
To evaluate the effectiveness of our semantics-oriented dataset expansion technique, we first isolate the contributions of its two main steps: data augmentation and data relabeling. We then compare our method with the widely used Face X-ray data augmentation~\cite{li2020face} and the human-in-the-loop data annotation in FFSC~\cite{zou2024semantic}. 

\begin{table}[!t]
\renewcommand{\arraystretch}{1.2}
\setlength{\abovecaptionskip}{0cm}
\caption{AUC results of SJEDD trained using different variants of our dataset expansion technique applied to FF++.  ``w/o relabeling'' corresponds to binary classification}
\label{tab:  sjedd_data_1}
\centering
\resizebox{\linewidth}{!}{
\begin{tabular}{ccccccc}
\toprule
Augmentation & Relabeling & CDF & DF-1.0 & DFDC & FFSC & Mean AUC  \\
\hline
\noalign{\smallskip}
\xmark & \xmark  & 81.45 & 86.78 & 76.00 & 78.04 & 80.57 \\
\xmark & \cmark  & 84.58 & 87.37 & 79.19 & 78.56 & 82.43 \\
\cmark & \xmark  & 86.39 & 91.07 & 81.39 & 82.52 & 85.34\\
\cmark & \cmark &  \textbf{92.22} & \textbf{93.21} & \textbf{84.14} & \textbf{85.70} & \textbf{88.82} \\
\bottomrule
\end{tabular}
}
\end{table}

As shown in Table~\ref{tab: sjedd_data_1}, data relabeling alone marginally improves SJEDD's generalization to unseen datasets. This happens because the relabeling process simply merges DF~\cite{faceswap} and FS~\cite{faceswap_Kowalski} into one category, and F2F~\cite{thies2016face2face} and NT~\cite{thies2019deferred} into another in FF++, which does not fully fulfill the potential of semantics-oriented categorization of face manipulations. As a result, SJEDD essentially reduces to a manipulation-oriented detector with a risk of overfitting.
Training with data augmentation consistently improves detection performance, aligning with the findings in \cite{li2020face, chen2022self, shiohara2022detecting}. The best performance is achieved when both data augmentation and relabeling are activated.
The improvements arise because our proposed semantics-oriented dataset expansion method enables a single manipulation to alter multiple attributes, and meanwhile groups multiple manipulations under the same attribute category. This encourages SJEDD to learn transferable features across different manipulations that affect similar face attributes, rather than relying solely on manipulation-specific cues.

\begin{table}[!t]
\renewcommand{\arraystretch}{1.2}
\setlength{\abovecaptionskip}{0cm}
\caption{AUC results of SJEDD variants trained using different data augmentation methods }
\label{tab:  sjedd_data_2}
\centering
\resizebox{\linewidth}{!}{
\begin{tabular}{lccccc}
\toprule
Training & FF++ & CDF & DF-1.0 & DFDC & FFSC   \\
\hline
\noalign{\smallskip}
Face X-ray-augmented FF++& \textbf{---} & 89.58 & 93.28 & 81.14 & 82.15 \\
Human-annotated FFSC  & 93.81 & 91.28 & \textbf{97.16} & 81.32 & \textbf{---} \\
\hline
Semantics-oriented FF++ (Ours)  & \textbf{---} & \textbf{92.22} & 93.21 & \textbf{84.14} & \textbf{85.70} \\
\bottomrule
\end{tabular}
}
\end{table}

As shown in Table~\ref{tab: sjedd_data_2}, the Face X-ray-trained SJEDD performs worse than the proposed SJEDD. Furthermore, our detector achieves results comparable to the FFSC-trained variant, which is exposed to a larger set of real-world sophisticated face manipulations and more comprehensive global face attributes. This underscores the effectiveness of our technique in automatically expanding existing DeepFake datasets. 

\subsection{Interpretability}
The proposed SJEDD via vision-language correspondence enhances model interpretability by providing human-understandable explanations. Specifically, we leverage SJEDD trained on FFSC~\cite{zou2024semantic} to illustrate this, as shown in Fig.~\ref{fig: visual_language_corres}. SJEDD accurately highlights the modified face attributes and regions with high probabilities. For example, in Fig.~\ref{fig: visual_language_corres}(e), the detector identifies the face as fake and meanwhile indicates a likely expression alteration, pinpointing the mouth and lip as manipulated regions. This allows us to combine the predefined text templates and form a textual explanation such as  ``\textit{The face image is fake because the global face attribute of expression is altered, through manipulation of the local lip and mouth regions}.'' As for images in FF++, CDF, and DFDC, SJEDD can still successfully interpret the binary classification results by predicting the authenticity of individual face attributes and regions. For example, on FF++, our detector accurately identifies DF and FS as instances of \texttt{identity} manipulation and F2F and NT as examples of \texttt{expression} forgery.

We additionally compare SJEDD with a baseline detector (\ie, ViT-B~\cite{dosovitskiy2020image} trained via binary classification on FFSC) using Grad-CAM visualizations~\cite{selvaraju2017grad}, where warmer regions are more important. As shown in Fig.~\ref{fig: attention_heatmap}, SJEDD provides more human-aligned visualizations by accurately attending to semantically meaningful (and manipulated) face regions.

\section{Conclusion and Discussion}\label{sec: conclusion}
We have described a semantics-oriented joint embedding DeepFake detection method called SJEDD, along with a dataset expansion technique. The core of
SJEDD lies in the use of joint embedding, which reformulates model parameter splitting in multitask learning to a model capacity allocation problem, allowing for end-to-end bi-level optimization. We replaced the cross-entropy loss with the fidelity loss in bi-level optimization, which additionally boosts detection performance. 

The present study addresses DeepFake detection purely from a technical standpoint. Future directions in this field are multifaceted, involving advancements in both technology and policy. 
From a technological perspective, it is recommended to further explore a multimodal approach that integrates audio, visual, and textual analysis, combined with advanced parameter-efficient fine-tuning techniques like LoRA~\cite{hu2022lora}, to improve training efficiency and detection accuracy.
Detectors should also be able to operate in real-time on edge devices to support live analysis of streaming content in large-scale media platforms. Another important avenue for advancement is to develop detectors capable of continual learning and adaptation to emerging DeepFake techniques, potentially involving human oversight to maintain detection accuracy and efficacy. Proactive measures such as blockchain technology and digital watermarking to authenticate original content should also be considered.
From a policy perspective, it is essential to establish international standards and frameworks for DeepFake detection. This would involve collaboration between governments, tech companies, and research institutions. Laws and regulations should also be enacted to address the creation and distribution of DeepFake content, including penalties for misuse and support for victims.

\section*{Acknowledgments}
This work was supported in part by the Hong Kong RGC General Research Fund (11220224), the CityU Strategic Research Grants (7005848 and 7005983), and an Industry Gift Fund (9229179).

\bibliographystyle{IEEEtran}
\bibliography{DeepFake}

\end{document}